\def\PublicVersion{1}
\documentclass{article}
\usepackage[T1]{fontenc}
\usepackage{iclr2027_conference,times}
\newif\ifarxivversion
\arxivversiontrue
\ifdefined\PublicVersion\arxivversiontrue\fi
\ifarxivversion
  \iclrfinalcopy
  \usepackage{etoolbox}
  \makeatletter
  \patchcmd{\@maketitle}{Published as a conference paper at ICLR 2027}{Preprint}{}%
    {\PackageError{publication-mode}{Could not replace the camera-ready header}{Check the ICLR style before publishing.}}
  \makeatother
\else
  \AtBeginDocument{\hypersetup{pdfauthor={}}}
\fi

\usepackage{amsmath,amssymb,amsthm}
\usepackage{mathtools}
\newcommand{\dff}{r}

\usepackage{graphicx}
\usepackage{booktabs}
\usepackage{array}
\usepackage{caption}
\usepackage{url}
\usepackage[colorlinks=true,linkcolor=blue,citecolor=blue,urlcolor=blue]{hyperref}
\usepackage{enumitem}
\usepackage{microtype}
\usepackage{needspace}
\usepackage{cleveref}

\usepackage{ifthen}  

\allowdisplaybreaks[1]

\newsavebox{\abortinterventionbox}

\newtheorem{theorem}{Theorem}

\newtheorem{proposition}[theorem]{Proposition}
\newtheorem*{thmrestate}{Theorem~\ref{thm:illegal}}

\newtheorem*{thmrestatew}{Theorem~\ref{thm:width}}
\theoremstyle{remark}
\newtheorem{remark}{Remark}

\newcommand{\xtok}{\texttt{x}}
\newcommand{\WMp}{W^{M'}}
\newcommand{\WVp}{W^{V'}}
\newcommand{\Atrue}{A^{\mathrm{true}}}
\newcommand{\Rtrue}{R^{\mathrm{true}}}
\newcommand{\Robs}{R^{\mathrm{obs}}}
\newcommand{\WA}{W^{A}}
\newcommand{\WU}{W^{U}}
\DeclareMathOperator*{\softmax}{softmax}

\newcommand{\compilehidecomments}{true}
\ifthenelse{ \equal{\compilehidecomments}{true} }{
	\newcommand{\siwei}[1]{}
	\newcommand{\wei}[1]{}
	\newcommand{\yifei}[1]{}
	\newcommand{\shi}[1]{}
	\newcommand{\shanghua}[1]{}
}{
	\newcommand{\siwei}[1]{{\color{red} [\text{Siwei:} #1]}}
	\newcommand{\wei}[1]{{\color{purple} [\text{Wei:} #1]}}
	\newcommand{\yifei}[1]{{\color{teal} [\text{Yifei:} #1]}}
	\newcommand{\shi}[1]{{\color{green} [\text{Shi:} #1]}}
	\newcommand{\shanghua}[1]{{\color{blue} [\text{Shanghua:} #1]}}
}

\title{
    Learning When to Abort in Transformer\\
    Planning: A Mechanistic Study}

\ifarxivversion% Author order and affiliations confirmed by the authors.
\author{\textbf{Shi Feng\textsuperscript{1} \quad Qimin Zhong\textsuperscript{2} \quad Siwei Wang\textsuperscript{3} \quad Yifei Shen\textsuperscript{3}}\\
\textbf{Shang-Hua Teng\textsuperscript{4} \quad Hao Liao\textsuperscript{2} \quad Wei Chen\textsuperscript{3}\thanks{Corresponding author: Wei Chen, \texttt{weic@microsoft.com}.}}\\[4pt]
{\normalfont\small \textsuperscript{1}Harvard University \quad \textsuperscript{2}Shenzhen University}\\
{\normalfont\small \textsuperscript{3}Microsoft Research Asia \quad \textsuperscript{4}University of Southern California}}
\AtBeginDocument{\hypersetup{pdfauthor={Shi Feng, Qimin Zhong, Siwei Wang, Yifei Shen, Shang-Hua Teng, Hao Liao, Wei Chen}}}
\fi

\begin{document}
\maketitle

\begin{abstract}

Planning does not by itself imply the ability to recognise errors, but recognising errors can improve planning capacity.
%
%, with our focus on the underlying mechanism.
%
In this paper, we study this phenomenon in Transformer planning.
%
%We study how supervision from such transitions can improve planning,
%with a focus on the underlying mechanism. 
%
To analyze the inner mechanism, we follow the approach of \cite{alpine} and abstract planning as path-finding in graphs. 
In this setting, we train the model to detect the error of invalid transitions, i.e., after the model generates a 
	next node that is not adjacent with the current node, the model should then generate a special abort token \xtok.
%
%
%Starting from correct trajectories, we
%construct error trajectories consisting of a valid prefix, an invalid transition,
%and a special abort token. Only the abort token receives loss on these examples.
%At inference, emitting this token causes an external procedure to discard the
%attempt and restart from the original prompt. In graph path-finding, 
For expressiveness, we construct a %one-layer, two-head 
Transformer with % embedding size $O(|V|)$ %and 
feed-forward width $O(|V|)$ 
for any fixed graph with node set $V$. It can detect all possible invalid transitions, and select a valid next node if the prefix is valid. %, after a valid transition, selects a valid next node when a valid continuation
%exists. 
We then propose a worst-case lower bound on feed-forward width, which makes the detector's linear width near-optimal up to
logarithmic factors. % within the specified class of rectified linear unit (ReLU)
%detectors. 
In trained models, we demonstrate that invalid-transition detection increases previous-node attention,
%interventions connect that information to the abort decision, and direct logit
%attribution 
and most of the valid--invalid separation locates in the feed-forward
network's output. 
This accords with our theoretical insights. 
%Adjacency probes also show slightly stronger edge--non-edge
%separation. 
Finally, we use this error detection to improve the model's performance: at inference, generating the abort token \xtok~will
directly cause an external procedure to discard the attempt and restart from the original prompt.
Experiments show that error detection combined with retry improves model's planning reliability on graphs, Countdown, and mazes.

\end{abstract}

\section{Introduction}
\label{sec:intro}

Planning involves choosing a sequence of steps that leads from an initial state
to a goal, and is ubiquitous in daily life.  In recent years, Transformer-based large language models (LLMs) are increasingly used for tasks requiring such multi-step planning, including mathematics and coding.
Understanding how these models plan, and improving their planning capacity therefore becomes more and more important.
%research
%is important for improving their reliability.
%
In this paper, we study how learning to recognise invalid planning steps (or invalid transitions), could improve language-model planning.
Traditional training on correct planning trajectories teaches a model which steps to take,
but does not explicitly teach how to recognise invalid steps.
In fact, planning ability does not need to bring failure recognition: a model may be able to find
valid solutions yet fail to recognise that a problematic input contains invalid steps.

In human planning, however, failure recognition is also thought to play an important role in planning ability \citep{alexander2010computational, fu2023neurophysiological}. %\siwei{Can we have some refs here?}\shi{added} Therefore, we focus on the following question:

\siwei{I revise the question, please double check.}
{\em Can a Transformer-based language model learn to detect invalid planning steps, what could be the underlying mechanism of this learning process, and how can this improve the model's performance?}

We conduct a mechanistic study on this problem, based on the path planning setup of ALPINE~\citep{alpine}.
In ALPINE, the planning task is abstracted as looking for valid paths in a fixed graph. %a Transformer is trained on source--target pairs and valid paths
%connecting them in a fixed graph. 
The graph is not supplied as an input, hence the model
must learn the graph's structure from the training paths and find valid paths for unseen pairs.
As for invalid planning steps, we focus on {\em invalid transition}, which is a step in the generated path from some node $u$ to node $v$ that has no corresponding edge in the underlying graph.
To enable the Transformer to detect invalid transitions, we generate new training data containing valid path prefixes followed by an invalid transition, and then followed by a
	special abort token $\xtok$.
Only the abort token is supervised on these generated data, so that the Transformer could learn to flag invalid transitions without being trained to produce them.
Intuitively, this is analogous to adding wrong planning steps and masking them in a planning task for the model to learn from mistakes. 

%At inference, a {\em harness}, an external procedure surrounding the model, discards
%the current attempt when $\xtok$ is emitted and restarts from the original prompt,
%giving the model another opportunity to find a valid path.

Graph path-finding lets us distinguish two computations. Selecting the next node
uses the current node's outgoing edges and reachability to the target; detecting
an invalid transition instead requires relating the previous node to the current one.

Learning from search traces and error markers can improve
planning~\citep{sos,searchformer,huangplanning,selfback}. Our focus is the mechanism:
what computation invalid-transition detection requires, and whether abort supervision
induces it in trained models.
%and inference
%\wei{Shi, can you fill out this part, introducing a paragraph or so on what studies are done in the technical sections? Such as Transformer construction, lower bound result,
%	investigation on attention and adjacency information (weight gap analysis). Not sure if you still have gradient-based analysis.
%} 
Theoretically, Proposition~\ref{thm:additive} of Section~\ref{sec:additive} shows that an additive score
of the previous and current nodes with a threshold cannot detect arbitrary invalid transitions.
In \Cref{sec.construction}, we construct a one-layer Transformer that combines
next-node selection with invalid-transition detection using the non-linear feed-forward layer with width
linear in the number of nodes. The lower bound in \Cref{sec.widthlowerbound}
makes the detector's linear width near-optimal up to logarithmic factors within
the specified class of ReLU detectors. Empirically, \Cref{sec:empirical} connects
the abort decision to previous-node information and feed-forward network (FFN)
processing in real trained Transformers.
Attention measurements show whether the model looks back, interventions test
whether previous-node information controls the abort decision, and direct logit
attribution measures the FFN's contribution to the abort signal.
%, and finds slightly stronger edge--non-edge separation in the learned
%adjacency probe. 
A fitted additive-logit alternative retains substantial detection
error on the same supervised prefixes (Appendix~\ref{app:additivefit}). These
findings are consistent with the computational dependencies identified by the theory.

We then ask whether the learned signal improves planning. At inference, a
{\em harness}, an external procedure surrounding the model, discards an attempt
when $\xtok$ is emitted and restarts from the original prompt.
Experiments on random graphs, Countdown and mazes show that this procedure improves
planning reliability (\Cref{sec:endtoend}). % and Appendix~\ref{app:additional}).
%\siwei{Here we may talk about best-of-N and our method's difference?}\shi{added}
Unlike verifier-ranked best-of-$N$ sampling~\citep{cobbe2021training,lightman2024let},
it does not require a fixed set of complete candidates or a separate correctness
verifier: the model supplies the signal for restarting an attempt.

Our contributions are threefold: a simple training method that turns local errors
into abort supervision; a mechanistic account of invalid-transition detection,
combining a one-layer Transformer construction, a capacity lower bound and
experiments on trained models; and an inference-time restart harness that uses
the learned signal to improve planning across tasks and models.

\section{Related Work}

Our mechanistic starting point is ALPINE~\citep{alpine}, which explains how
Transformers select the next node using adjacency and reachability information.
We build on this account to study invalid-transition detection.
Complementary work studies how models compose reachability information to select
future steps~\citep{mtpplan}, represent and maintain the current planning state
during search~\citep{verifysearch},
and improve planning through reinforcement-learning fine-tuning~\citep{wang2026benefits}.

Our training approach belongs to the line of work on learning from search and
error traces. \citet{sos} train on complete search trajectories containing
exploratory mistakes and backtracking, while \citet{searchformer} train models to
imitate A$^*$ search dynamics. Closer to our setting, \citet{huangplanning} train
on erroneous actions followed by in-context corrections. Their models recognise
mistakes, but self-correction does not improve plan validity in their evaluation.
We instead separate detection from recovery: the model learns to flag an invalid
transition, and an external procedure handles retry.

Self-Backtracking~\citep{selfback} is the closest method. It learns a token that
triggers rollback and further search from an earlier state; our harness restarts
from the original prompt. Its error-example loss supervises both the correct
prefix and the backtrack token, whereas ours supervises only the abort token.
Our labels require checking one transition, rather than determining that a prefix
admits no valid completion, which can require searching over future
continuations~\citep{bylander1994computational}. In our small-model Countdown
experiments, we cross the two objectives with restart and rollback decoders.
Abort-only supervision performs better with either decoder, suggesting that the
simpler objective contributes to the gain in this setting
(Appendix~\ref{app:selfback}). Our mechanistic analysis further distinguishes this
work by combining an explicit Transformer construction with experiments on the
learned computation.

Additionally, some work uses separate signals to guide generation. State--action
verifiers~\citep{aroraverifier} check whether an action is valid in the current state.
Process-verifier approaches use information about whether a prefix admits a valid
completion~\citep{verifierquery}, or learned estimates of the quality of partial
generations~\citep{taming}, to guide sampling and backtracking. In our approach,
the planner itself (rather than an independent verifier) learns to emit the abort signal during generation.
Confidence-token methods~\citep{conftok} are also related, but estimate the reliability
of the overall answer rather than whether the latest step is invalid.

\section{Model and Method}
\label{sec:setting}

\subsection{Transformer Architecture}
\label{sec:architecture}

We use the following one-layer architecture for the theoretical analysis, following
\citet{alpine}.
Let $N$ be the sequence length, $d$ the
embedding size, $H$ the number of attention heads, $d_k = d/H$ the size of each head,
and $M$ the vocabulary size. A token sequence is a one-hot matrix
$U \in \mathbb{R}^{N\times M}$, embedded as $H_0 = U W_t + W_p$, where
$W_t\in\mathbb{R}^{M\times d}$ and $W_p\in\mathbb{R}^{N\times d}$ are the token and
positional embedding matrices. The multi-head
attention (MHA) sublayer is
\begin{equation}
\mathrm{MHA}(H_0) = \Big[\softmax\Big(\tfrac{(H_0W_Q^{(h)})(H_0W_K^{(h)})^\top}{\sqrt{d_k}}\Big) H_0 W_V^{(h)}\Big]_{h=1}^{H} \WA ,
\end{equation}
where the softmax is causally masked and $W_Q^{(h)}$, $W_K^{(h)}$, and $W_V^{(h)}$ are
the $d\times d_k$ query, key, and value projections for head $h$, and the brackets denote
horizontal concatenation over heads. The attention output projection is
$\WA \in \mathbb{R}^{d\times d}$. With $\tilde H = \mathrm{MHA}(H_0) + H_0$, the feed-forward network is
$\mathrm{FFN}(\tilde H) = \mathrm{ReLU}(\tilde H W_1 + b_1)W_2 + b_2$, where
$\mathrm{ReLU}(z)=\max(0,z)$ acts elementwise, and the layer output is
$H_1 = \mathrm{FFN}(\tilde H) + \tilde H$. Here
$W_1\in\mathbb{R}^{d\times\dff}$, $b_1\in\mathbb{R}^{\dff}$,
$W_2\in\mathbb{R}^{\dff\times d}$, and $b_2\in\mathbb{R}^{d}$. The feed-forward width $\dff$ is
conventionally fixed at $4d$. We name $\dff$ explicitly because it is the resource the
abort detector uses. The vocabulary logits (next-token scores before softmax) are $H_1\WU$, where
$\WU \in \mathbb{R}^{d\times M}$ is the unembedding matrix. Layer
normalisation is omitted in the above equations.

\subsection{Path-Planning Task, Data, and Training}

\siwei{Maybe we should introduce all the three datasets here. }\shi{I think it is okay to introduce other datasets in section 6 since section 4, 5 are both specific to graph path planning.}

We follow the graph path-finding setup of ALPINE~\citep{alpine}. Fix a directed acyclic
graph (DAG) $G=(V,E)$ with $n=|V|$ nodes. Given a source $s$ and a reachable target $t$,
the model receives the prompt \texttt{s t} and needs to generate a node sequence starting at $s$,
ending at $t$, and following graph edges. We define a \emph{valid path prefix} as a node sequence that starts at $s$ and follows only graph edges, and write $u\rightsquigarrow t$ when $u$ can reach $t$, including $t\rightsquigarrow t$.

There are two types of training trajectories. Correct training trajectories have the form \texttt{s t s $u_1$ \dots\ t}, where \texttt{s $u_1$ \dots\ t} is a valid path from $s$ to $t$. These trajectories are used to train the usual sequence generation process, and thus receive the usual next-token loss. Error trajectories add supervision about invalid transitions. Each contains a valid prefix followed by one
non-edge $p\to i$ and the token \xtok. Only the final \xtok\ is supervised;
the prefix and invalid transition provide context, not prediction targets.
We call this training signal \emph{abort supervision}.
We call the resulting model a $+\xtok$ model, and a model trained only on correct
trajectories the \emph{baseline}. The detailed loss formulation is given in \eqref{eq:loss} (Appendix \ref{app:setting}).

%and receive
%the usual next-token loss. Error trajectories add supervision about invalid transitions.
%Each contains a valid prefix followed by one
%non-edge $p\to i$ and the token \xtok. Only the final \xtok\ is supervised: the prefix
%and incorrect step are context, not prediction targets. We call the resulting model a $+\xtok$ model, and a model trained only on correct
%trajectories the \emph{baseline}. The loss is given in \eqref{eq:loss}.

%Correct training trajectories have the form \texttt{s t s $u_1$ \dots\ t} and receive
%the usual next-token loss. Error trajectories add supervision about invalid transitions.
%Each contains a valid prefix followed by one
%non-edge $p\to i$ and the token \xtok. Only the final \xtok\ is supervised: the prefix
%and incorrect step are context, not prediction targets. We call the resulting model a $+\xtok$ model, and a model trained only on correct
%trajectories the \emph{baseline}. The loss is given in \eqref{eq:loss}.

Our abort labels are local: in graph planning, determining whether a transition
$p\to i$ is invalid only requires checking whether $(p,i)\in E$, without reasoning
about whether the current prefix can still reach the target.
%\siwei{Not very sure what you mean in the above sentence...}
%\shi{I revised the sentence.}
%\siwei{I just feel we do not need to explain this? In fact, I think what you want to present is that this is a local check and does not require much capacity. In this case, we may highlight more on the local part, instead of the difference between dead end and invalid transition.}\shi{fixed}
Thus, constructing error trajectories requires only a local edge check, rather than
solving the remaining planning problem.

For graph experiments, each edge $i\to j$ with $i<j$ is included independently with
probability $0.1$. Reachable source--target pairs are randomly split into training
and test sets, except that direct-edge pairs always enter training. Each training
pair contributes $20$ valid paths. The $+\xtok$ dataset adds one error trajectory
per correct trajectory, and minibatches sample uniformly from this mixture.
The loss averages over supervised tokens: correct trajectories supply multiple
labels, error trajectories only one. Thus, correct trajectories make up half
the batch on average but supply most labels. %The control in
%Appendix~\ref{app:budget} doubles $+\xtok$ updates to match the baseline's expected
%number of correct trajectories seen.
%\siwei{Is it really a half? Do you reweight the loss in different trajectories?}\shi{The ``half'' refers to the number of correct trajectories sampled, not the token-level loss weight. We do not reweight trajectory types; the loss is averaged over supervised tokens. I revised the text to make this distinction explicit.}

We denote an architecture by layer count/head count/embedding size and compare
$1/1/120$, $1/2/120$ and $1/2/384$. All use feed-forward width $\dff=4d$, where $d$ is
embedding size. The experiments use GELUs (Gaussian error linear units), layer
normalisation and tied embedding/unembedding weights.
%\siwei{Do we need to introduce some about transformers before?}\shi{I added section 3.1.}
Models are trained for $10^4$ updates with batch size
$1{,}024$ and learning rate $5\times10^{-4}$; Appendix~\ref{app:setting} gives the full
protocol. Graph results use five seeds, each with an independently
generated graph. Error bars are standard errors (s.e.) over seeds and gains are paired.
% Editorial note: if we introduce all the three datasets here in this section, we also need to include some details for the other two datasets.

\section{Theory of Invalid-Transition Detection in Path-Planning}
\label{sec:theory}

Let $\Atrue_{(p,i)}=\mathbf{1}[(p,i)\in E]$ be the graph's adjacency matrix, where
$\mathbf{1}[\cdot]$ is $1$ when its condition holds and $0$ otherwise.
The true reachability matrix has entry $\Rtrue_{(t,u)}=\mathbf{1}[u\rightsquigarrow t]$.
%\siwei{Not very clear here. Do you mean the next token should depend on $(i,t,p)$? Is it an assumption, a lemma, or just an intuition?}\shi{I deleted it because it is the condition of next subsection.}

\subsection{Additive Detection Is Insufficient}
\label{sec:additive}

To output the abort token correctly, the model needs to determine whether the most recent
transition $p\to i$ is valid, that is, whether $(p,i)\in E$. We next show that simply adding
separate functions of $p$ and $i$ in a single thresholded score is not sufficient
for general graphs.
%\siwei{The score is a little sudden, and I do not really understand this sentence. If you want to say "additive score is not enough", I suggest to write Theorem 1 at the beginning of this subsection. Below can be a simple proof sketch. Also, the "Interaction" in the title is confusing. }\shi{You are right! I changed the sequence.}

\begin{proposition}
\label{thm:additive}
If $\Atrue$ has nodes $p\ne p'$ and $i\ne i'$ with
$\Atrue_{(p,i)}=\Atrue_{(p',i')}=1$ and
$\Atrue_{(p,i')}=\Atrue_{(p',i)}=0$,
then no threshold on an additive function $f(p)+g(i)$ can correctly classify all four
transitions as edges or non-edges, whether larger scores indicate validity or invalidity.
\end{proposition}

To see why, consider the two edges $(p,i)$ and $(p',i')$, whose crossed pairs
$(p,i')$ and $(p',i)$ are non-edges. For an additive function
$\phi(p,i)=f(p)+g(i)$, with $f,g:V\to\mathbb{R}$,
\[
\phi(p,i)+\phi(p',i')
=
\phi(p,i')+\phi(p',i).
\]
Thus, the two edge scores and the two non-edge scores have the same total, so no threshold
can place both edges on one side and both non-edges on the other.

%\siwei{I think there should be a consistent name for our detection. "Invalid transition detection" or "adjacency detection"? Which one do you prefer? Or any other choices?}
%\shi{We should use invalid transition detection.}

Thus, general invalid-transition detection with a scalar threshold requires a
non-additive combination of previous-node and current-node information. In the next
subsection, we show how a Transformer implements this check by selecting the
current node's entry from the previous node's adjacency row.

%construction implements it as a lookup of the current node in the
%previous node's adjacency row.

\subsection{A Gated Lookup for Invalid Transition Detection}
\label{sec.construction}

%\siwei{Similarly, the "Interaction" in the title is confusing. Also, without a introduction for transformer models, I am not sure whether we can convince the readers about the correctness of construction. For example, the paragraph below may be a little confusing...}\shi{I removed "interaction" and we have the transformer definition now.}

In this subsection, we construct a one-layer, two-head Transformer that performs path planning and invalid-transition detection simultaneously. We consider input sequences of the form ``$s\ t\ s\ u_1 \dots u_k$'', where $s\rightsquigarrow t$.

\begin{theorem}
\label{thm:illegal}
Given a DAG $G$ with adjacency matrix $\Atrue$ and reachability matrix $\Rtrue$, for
every $\varepsilon > 0$ there is a $1$-layer, $2$-head Transformer with embedding size
$O(|V|)$ and feed-forward width $O(|V|)$ that, for every input sequence
``$s\ t\ s\ u_1 \dots u_k$'' for which $(u_0,\dots,u_{k-1})$ is a valid path prefix,
where $u_0=s$, $k \geq 1$, and $u_k \neq t$, generates with probability
at least $1-\varepsilon$
\begin{enumerate}[label=(\roman*),leftmargin=*,itemsep=1pt]
  \item the token \xtok, if $\Atrue_{(u_{k-1},\,u_k)} = 0$, and
  \item otherwise a node $u_{k+1}$ that is both an out-neighbour of $u_k$ and able to
        reach $t$, whenever such a node exists.
\end{enumerate}
\end{theorem}

\siwei{I try to revise this proof sketch, but I think there are too many things unclear and perhaps required a major revision. Please double check it. Also, in the last section you said the non-linear part is GELU, but here is RELU. Is that also a problem?} \shi{Checked and GELU is indeed different. We may say we have RELU as simplified version here.}

Write $p=u_{k-1}$ and $i=u_k$ for the previous and current nodes. One attention
head retrieves the target's reachability row $\Rtrue_{(t,:)}$; the other retrieves
the previous node's adjacency row $\Atrue_{(p,:)}$. The residual connection retains
the current node's one-hot vector $e_i$. Thus, at exact attention the relevant FFN
input blocks are $\Rtrue_{(t,:)}\oplus\Atrue_{(p,:)}\oplus e_i$, where $\oplus$
denotes concatenation.

The FFN's first linear map reads the current node's adjacency row
$\Atrue_{(i,:)}$ from $e_i$, and thus the parameter becomes $\Rtrue_{(t,:)}\oplus\Atrue_{(p,:)}\oplus e_i \oplus \Atrue_{(i,:)}$. Its first $|V|$ ReLU units use $\Rtrue_{(t,:)}$ and $\Atrue_{(i,:)}$, compute
$\max(0,\Rtrue_{(t,v)}+\Atrue_{(i,v)})=\Rtrue_{(t,v)} + \Atrue_{(i,v)}$,
selecting out-neighbours of $i$ that can reach $t$ (if such nodes exist), as in the adjacency--reachability
account of \citet{alpine}.
%\siwei{What if there is no feasible solution? For example, $i \not\rightsquigarrow t$?}
%\shi{The next-node guarantee requires a valid continuation. The detector still checks the preceding transition when none exists; its abort logit does not depend on reachability.}
The other $|V|$ units combine $\Atrue_{(p,:)}$ with $e_i$ to check the preceding
transition. For each node $v$, a unit computes
\begin{equation}
\max\big(0,\mathbf{1}[i=v]+\Atrue_{(p,v)}-1\big)
=\mathbf{1}[i=v]\,\Atrue_{(p,v)}.
\label{eq:gate}
\end{equation}

%The detection units sum to $\Atrue_{(p,i)}$. 
Their output suppresses the abort logit when the preceding transition is valid (when the sum of detection units is 1), and otherwise do nothing, letting the abort logit be largest (when the sum of detection units is 0). %After a valid transition,
%out-neighbours that reach $t$ have the largest logits whenever such nodes exist.

%One attention head supplies the target's reachability row $\Rtrue_{(t,:)}$; the other
%supplies the previous node's adjacency row $\Atrue_{(p,:)}$. The residual stream supplies
%the current node $i=u_k$. For each node $v$, a ReLU unit computes the conjunction
%\begin{equation}
%\max\big(0,\mathbf{1}[i=v]+\Atrue_{(p,v)}-1\big)
%=\mathbf{1}[i=v]\,\Atrue_{(p,v)}.
%\label{eq:gate}
%\end{equation}
%Thus a lookup unit fires exactly when the last transition is valid. When it fires, it
%supplies the current node's adjacency row and suppresses the abort logit; combining
%adjacency with reachability makes a valid next node win. Otherwise the abort logit wins.
The detector uses $|V|$ units, while the complete FFN has width $2|V|$.
Appendix~\ref{app:illegal} gives the detailed architecture and weights and shows
that the logit separation persists with finite attention scores.

\begin{remark}[Observed reachability and retry]
\label{rem:observed-retry}
Let $\Robs_{(t,u)}=1$ when a training path ending at $t$ contains $u$.
Using $\Robs$ instead of $\Rtrue$ can affect next-node selection, but not
invalid-transition detection. %, which depends only on whether the preceding step
%is an edge. 
%
%
%
%With our local check, this generation will be aborted and the model needs to regenerate a new one. 
%
%
With exact checks of every transition (including the final one) and both endpoints,
every accepted path is valid without a global ground-truth check.
%
%
%Every accepted output is then valid,
If independent, length-bounded attempts each succeed with fixed probability
$\pi>0$, retry succeeds within $K$ attempts with probability $1-(1-\pi)^K$
and terminates almost surely without a fixed budget. Thus, retry can overcome
incomplete observed reachability~\citep{alpine}.
%If an attempt
%returns a valid path with probability $\pi>0$, rejecting all other attempts and
%retrying succeeds within $K$ attempts with probability $1-(1-\pi)^K$, which
%approaches $1$ as $K$ grows. 
\end{remark}
% \siwei{Is this a good solution? In my proof sketch above, I think it is OK to let the abort token have very large weight in the first part, and a larges minus weight in the second part (if one unit equal to 1). In that case, it can just detect invalid transition.}

\subsection{How Much Capacity Does the Detector Require?}
\label{sec.widthlowerbound}

Suppose that the previous node $p$, current node $i$, and target node $t$ are
represented at the FFN input by separate one-hot blocks, i.e., 
$x_{p,i,t}=e_p\oplus e_i\oplus e_t$. A width-$r$ ReLU FFN computes
$\mathrm{FFN}(x)=\mathrm{ReLU}(xW_1+b_1)W_2+b_2$.
Let $W_1^{(p)},W_1^{(i)},W_1^{(t)}\in\mathbb R^{n\times r}$ be the row blocks of
$W_1$ acting on the three node inputs. The $h$-th FFN unit receives
\[
W_1^{(p)}[p,h]+W_1^{(i)}[i,h]+W_1^{(t)}[t,h]+b_1[h].
\]
The FFN's contribution to the abort logit is
\begin{align}
z_{\xtok}^{\mathrm{FFN}}(p,i,t)
=
\bigl(\mathrm{ReLU}(x_{p,i,t}W_1+b_1)W_2+b_2\bigr)
\WU[:,\xtok].
\label{eq:FFN}
\end{align}
Since transition validity depends only on $(p,i)$, we can fix any target $t_0$.
Then $W_1^{(t)}[t_0,:]$ is constant and can be absorbed into $b_1$.
Thus, for a fixed target, invalid-transition detection reduces to a
width-$r$ ReLU network over the previous and current nodes.

\begin{theorem}
\label{thm:width}
Suppose that a width-$r$ one-hidden-layer ReLU FFN receives the previous node
$p$, current node $i$, and target $t$ as the input
$e_p\oplus e_i\oplus e_t$. If, for every DAG on $|V|$ nodes, there exist FFN
parameters for which the sign of $z_{\xtok}^{\mathrm{FFN}}(p,i,t)$ correctly
determines whether $(p,i)\in E$ for every $p,i,t\in V$, then
\[
r=\Omega\left(\frac{|V|}{\log^2 |V|}\right).
\]
\end{theorem}
\shi{I am not sure whether this is clear. I believe it is better than the previous version.}

To prove the theorem, %we fix the target $t$, reducing the problem to a ReLU
%network over the previous and current nodes. W
we bound the number of adjacency
matrices representable by a width-$r$ network using the VC dimension of
piecewise-linear networks~\citep{vcrelu} and the Sauer--Shelah lemma, and
compare it with the $2^{\binom{|V|}{2}}$ DAGs obtained by choosing arbitrary
forward edges under a fixed node ordering (Appendix~\ref{app:width}).
The detector in Theorem~\ref{thm:illegal} uses $|V|$ ReLU units, matching this
worst-case lower bound up to logarithmic factors in this class. %: the linear map
%$e_p\mapsto\Atrue_{(p,:)}$ can be absorbed into the detector's first linear layer
%without adding ReLU units.
%\siwei{Why these two appendices have the same hyperlink?}

\section{Mechanistic Analysis of Abort Supervision}
\label{sec:empirical}

In this section, we examine whether trained models implement the invalid-transition detection
mechanism suggested by our theoretical analysis. We study previous-node
attention and its causal role in the abort decision, then use direct logit
attribution (DLA) to locate the abort signal. Additional analyses of learned
adjacency and reachability representations are in Appendix~\ref{sec:planning-probes}.
%
%first analyze how the model implement invalid transition check by
%
%tracing previous-node information from attention to the FFN's contribution to the abort logit, test an additive alternative, and then examine rejection during generation.
%
We compare baseline and $+\xtok$ models with the 1/2/120 and 1/2/384
architectures. For each graph size and architecture, we train both conditions
from scratch on five independently generated graphs (Section~\ref{sec:setting}).

At the beginning, we check that these models can both find valid paths and recognise failures.
The 1/2/384 $+\xtok$ models produce a valid path in more than $90\%$ of single
draws at every tested graph size, averaging over the five graphs at each size
(Figure~\ref{fig:gain}a). Their abort precision, the fraction of emitted aborts
whose prefixes admit no valid completion, is $94$--$99\%$ across these sizes
(Appendix~\ref{sec:setup}).
%\siwei{The above part should be in the previous section. }
%
%
For the causal interventions and DLA, we use six independently generated
$n=300$ graphs, each with a trained 1/2/384 $+\xtok$ model. On held-out prefixes
with an injected invalid transition, these models assign a mean probability of
$99.7\%$ to the abort token (Appendix~\ref{app:causal}). This strong abort
response gives us a clear setting in which to study how the decision is made.

\subsection{Previous-Node Information Drives the Abort Decision}
\label{sec:previous-attention}

% \siwei{Either move the figure here, or we can just remove the sentence below.}
% Average attention maps show target- and current-position heads (Figure~\ref{fig:heads}).
% These average roles coexist with input-dependent attention to the previous node.

\begin{figure}[!htbp]
\centering
% Measure the table once so the plot and table have exactly the same height.
\begin{lrbox}{\abortinterventionbox}
\begin{minipage}{0.59\textwidth}
\centering
\small
\setlength{\tabcolsep}{3pt}
\renewcommand{\arraystretch}{1.3}
\begin{tabular}{@{}>{\raggedright\arraybackslash}p{0.60\linewidth}r@{}}
\toprule
Intervention & $P(\xtok)$ \\
\midrule
None & $0.987\pm0.003$ \\
Remove previous-node attention (main head) & $0.089\pm0.086$ \\
Remove previous-node attention (both heads) & $0.008\pm0.002$ \\
Swap previous node (valid prefix) & $0.001\pm0.000$ \\
Remove FFN output & $0.000\pm0.000$ \\
\bottomrule
\end{tabular}
\end{minipage}
\end{lrbox}
\begin{minipage}[t]{0.38\textwidth}
\vspace{0pt}
\captionsetup{type=figure}
\centering
\includegraphics[height=\dimexpr\ht\abortinterventionbox+\dp\abortinterventionbox\relax]{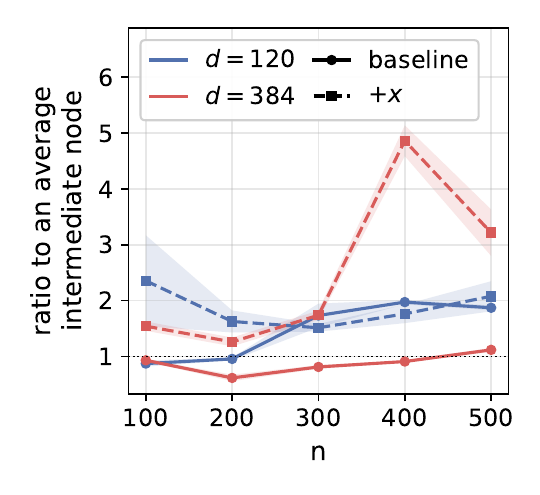}
\caption{Previous-node attention ratio over five seeds; shaded bands are s.e.}
\label{fig:mech}
\end{minipage}\hfill
\begin{minipage}[t]{0.59\textwidth}
\vspace{0pt}
\captionsetup{type=table}
\centering
\usebox{\abortinterventionbox}
\caption{Interventions after an injected invalid transition, on prefixes whose current node
can still reach the target. Mean $P(\xtok)$ $\pm$ s.e.\ over six $n=300$ graphs.}
\label{tab:causal-main}
\end{minipage}
\end{figure}

%\siwei{Here lacks explanation about how this attention be computed, e.g., over which kinds of data?}
We first test whether abort supervision increases attention to the previous node.
For each model, we feed 3,000
correct held-out paths as fixed input sequences and average the attention weights over
the two heads and over sequences containing each position. %Using zero-based token
%positions, 
We measure attention on the previous node (key $q-1$) and the mean
attention on the other intermediate emitted nodes (keys $3,\ldots,q-2$).
We average each quantity over $q=5,\ldots,8$, where this comparison set is
nonempty, and take their ratio. A ratio of $1$ means that the previous node
receives the same attention as an average intermediate node. This ratio rises from
0.88 to 2.52 at $d=384$ (Figure~\ref{fig:mech}), with an average paired increase of
$1.65\pm0.09$ over five seeds and five graph sizes. At $d=120$, the corresponding
increase is $0.39\pm0.25$. %\siwei{Does this mean the attention itself contain some capacity to do the invalid transition check? If this is what you want to represent, say it explicitly. }\shi{No. This measurement uses correct paths and only shows a general tendency to attend more to the previous node. The interventions below test whether this information actually affects the abort decision.}

\Needspace{7\baselineskip}
\begin{samepage}
The above measurement uses only correct paths and therefore captures the model's
general tendency to look back. We then test whether attention depends on
transition validity. On six $n=300$ graphs, the head with larger
previous-node attention on injected invalid prefixes assigns 0.58 attention
mass to the previous node after an invalid transition, compared with 0.17
after a valid transition.\par
\end{samepage}

\Needspace{10\baselineskip}
We next test whether previous-node information causally affects the abort decision.
We use the six $n=300$, 1/2/384 $+\xtok$ models described above.
For each graph, we take 2,000 held-out correct paths, cut each path at a
random interior node $u_k$, and append a node $w$ that is not a successor of $u_k$.
This produces an invalid prefix ``$s\ t\ u_0\ldots u_k\ w$'', and we measure
$P(\xtok)$ at the position immediately after $w$. %\siwei{A description about this prefix?}\shi{added}
We retain prefixes for which a previous-node replacement can make both adjacent
transitions valid, so the interventions use the same set of prefixes.

Table~\ref{tab:causal-main} reports interventions on the subset of prefixes for
which $w$ can still reach the target. For each graph, we call the head with more
attention to the previous node on invalid prefixes the main head. Removing
its attention to the previous node and renormalising lowers $P(\xtok)$ from
$0.99$ to $0.09$, while removing previous-node attention from both heads lowers
it to $0.008$. %\siwei{I add a line for removing main head in the table. Please fill the s.e.}\shi{added}%\siwei{which head?}\shi{defined}
We also change only the previous node. Specifically, we replace $u_k$ by a node
$p'$ such that $(u_{k-1},p')\in E$ and $(p',w)\in E$. The resulting prefix is
valid and differs from the injected-invalid prefix only at the previous-node
position. This change lowers $P(\xtok)$ to $0.001$.
Finally, we set the FFN output at the final position to zero before the
residual addition, and then apply the final layer normalisation and
unembedding as usual. This totally removes the abort response. %\siwei{How to drop this output?}

%These
%interventions (Table~\ref{tab:causal-main}) trace the abort response through previous-node
%attention and the FFN.
%\siwei{Do not understand the next sentence, and its relation between Table 1. Also, I believe this table is more likely to be in the subsection "Abort Supervision Makes the Model Look Back".}
All rows of Table~\ref{tab:causal-main} are restricted to prefixes whose injected
node $w$ can still reach the target. This avoids mixing invalid-transition
detection with dead-end detection: after the previous-node swap, the prefix is
valid and the target is still reachable, so the model no longer has a reason to
abort. Appendix~\ref{app:causal} gives the full protocol, controls, and
uncertainty estimates.

\subsection{Direct Logit Attribution Locates the Abort Signal in the FFN}
\label{sec:main-dla}

The interventions above show that previous-node information causally affects the
abort decision. We next take a closer look at where this information is converted into an abort signal, using direct logit attribution (DLA). Let $X$ be the sum of token and
positional embeddings. For our one-layer Transformer, the residual stream before
the final normalisation is
\[
H =
X+\operatorname{MHA}\bigl(\operatorname{LN}_1(X)\bigr)
+
\operatorname{FFN}\!\left(
\operatorname{LN}_2\!\left(
X+\operatorname{MHA}\bigl(\operatorname{LN}_1(X)\bigr)
\right)\right).
\]
The first two terms contain the information retrieved by attention, while the last
term is the output after this information passes through the nonlinear FFN.
Here $\operatorname{LN}_1$ and $\operatorname{LN}_2$ are the layer normalisations
before attention and before the FFN, respectively.

DLA measures how much each part directly contributes to the abort logit in the
same forward pass. For example, let $M$ denote the FFN output. Its contribution is
$z_{\xtok}^{M}=w_{\xtok}^{\top}[g\odot(M-\operatorname{mean}(M))/\operatorname{std}(H)]$.
All quantities here are evaluated at the final position: $w_{\xtok}$ is the abort
unembedding vector, $g$ is the final layer-normalisation scale vector, and
$\operatorname{std}(H)$ includes the numerical stabiliser. 
%
%In this experiment, we still use the paired invalid and swap-valid prefixes described in the previous section. 
%
%
%have the same final token $w$ and
%position, so their embedding vectors at that position are identical. Thus, the
%change in the pre-FFN representation at that position comes entirely from the
%attention output.
%
%Each prefix uses the
%normalisation scale from its own forward pass, so identical embedding vectors
%need not have identical normalised logit contributions across the pair.
Intuitively, this is the amount of abort logit that disappears if the FFN
contribution is set to zero.
%in this decomposition, while keeping the final
%normalisation scale from the original forward pass fixed. 
It is an attribution,
rather than rerunning the model with the FFN removed. 
We similarly compute the
contribution before the FFN. Comparing these contributions between valid and
invalid prefixes (as described in the last subsection) shows whether most of the valid--invalid separation is already
written directly before the FFN or instead appears in the FFN output.
%
%(Appendix~\ref{app:dla}).

%The paired invalid and swap-valid prefixes have the same final token $w$ and
%position, so their embedding vectors at that position are identical. Thus, the
%change in the pre-FFN representation at that position comes entirely from the
%attention output.
%\siwei{First, I think this method requires more explanation. I do not really understand why it answers the question "Where is this information converted into an abort score?" also, not very sure what does the following paragraph mean...}

On target-reachable invalid prefixes, the direct contribution before the FFN
changes little when the previous node is swapped to make the last transition valid
($-0.68$ to $-1.15$), while the FFN contribution drops from $+29.6$ to $+1.5$;
the total abort logit falls from $29.0$ to $0.3$. Thus, little of the
valid--invalid difference is written directly into the abort logit before the FFN;
most of it appears in the FFN output. The same pattern holds for the abort margin,
defined as the abort logit minus the largest node logit: the pre-FFN contribution
changes from $-1.0$ to $-1.7$, while the FFN contribution changes from $+22.4$
to $-10.2$. Since the FFN receives only the pre-FFN residual stream, the information
needed to distinguish the two prefixes must already be present there, but it is the FFN
transforms it into a strong abort-versus-continue signal. This accords with our analysis, i.e., an additive score is not enough ($X+\operatorname{MHA}\bigl(\operatorname{LN}_1(X)\bigr)$), and the ReLU in FFN layer is critical for the abort signal. Appendix~\ref{app:dla}
gives the full decomposition, including the unit-level analysis.

\section{Planning with the Learned Abort Signal}
\label{sec:endtoend}

We now ask whether the learned abort signal improves planning. We evaluate graphs,
Countdown, and mazes; Countdown and maze experiments use $1/2/384$ models. All
models are trained for $10^4$ updates with batch size $1{,}024$ and learning rate
$5\times10^{-4}$. As in Section~\ref{sec:setting}, correct trajectories receive
standard next-token supervision, while error trajectories supervise only the final
abort token. All results use sampling at temperature $1$.

We use a simple restart harness with draw budget $K$. If the $+\xtok$ model emits
$\xtok$, the current draw is rejected and generation restarts from the original
prompt. Otherwise, the completed output is checked by a domain-specific acceptance
test $\mathcal A$, which provides a second reason to reject a draw without checking
full correctness. In graphs, $\mathcal A$ requires at least two path-node tokens
and a final node equal to the target, but does not check intermediate edges. In
Countdown, it requires a nonempty, well-formed sequence of equations whose last
stated result equals the target, without checking arithmetic or operand availability.
In mazes, it accepts every non-aborting output. After any rejection, the model draws
again until an output is accepted or the budget $K$ is exhausted. The baseline uses
the same acceptance rule and budget, but has no learned abort signal.

We report three metrics: single-shot validity is the valid-output rate of one draw;
harness validity is the rate at which the harness returns a valid output, counting
exhausted budgets as failures; and mean draw count is averaged over all evaluation
instances, with an exhausted budget counted as $K$ draws
(Appendix~\ref{sec:setup}).
\siwei{Why letting it conditioned on acceptance? Even not accepted, it is still cost the budget, and in my understanding, we should count it as one generation.}
\shi{I changed it to average over all instances, with an exhausted budget counted as $K$ draws.}

\subsection{Graph Path-Finding}

%We use $K=100$ and $3{,}000$ held-out evaluation instances.

\begin{figure}[!htbp]
\centering
\includegraphics[width=0.7\textwidth]{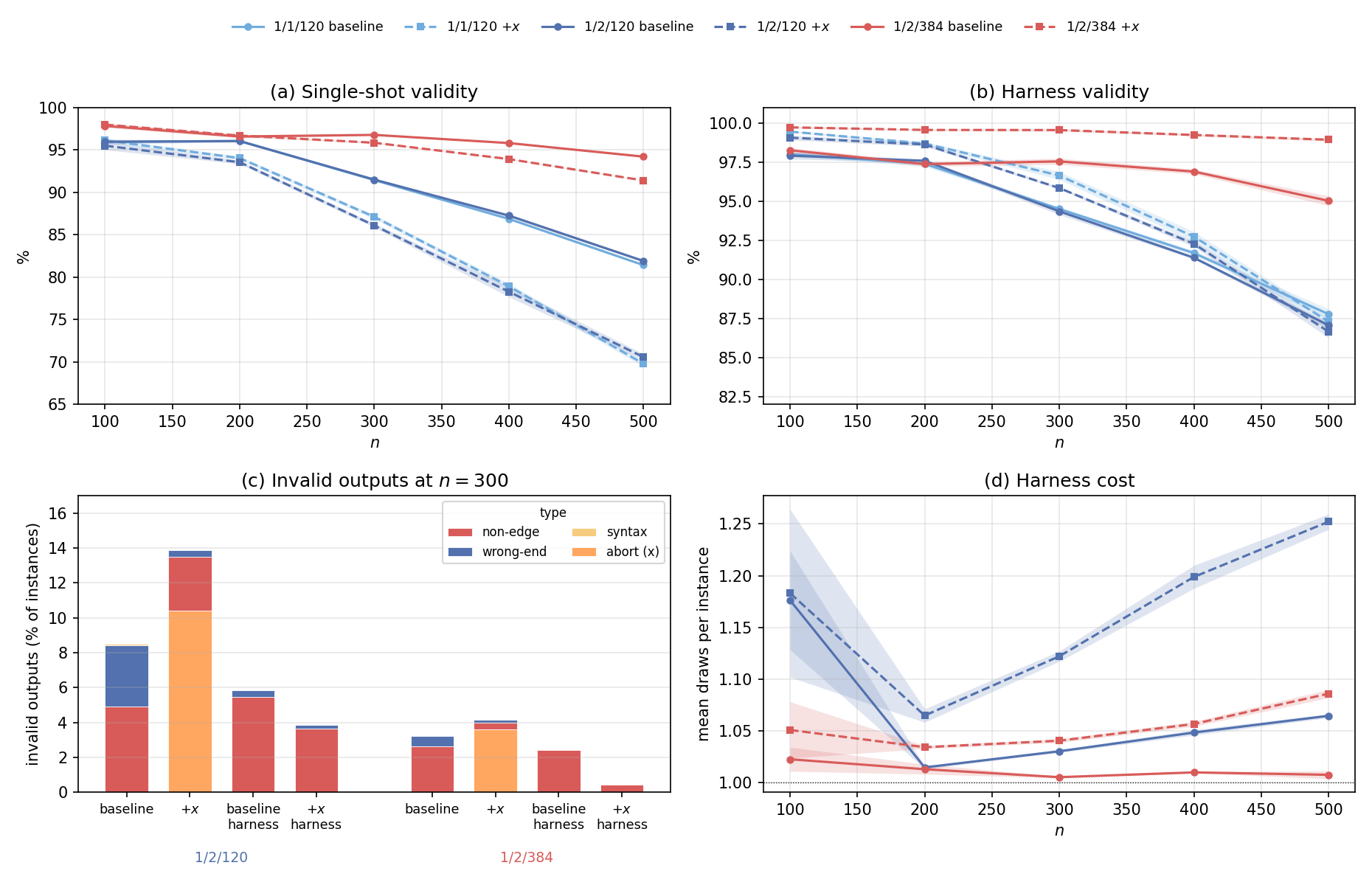}
\caption{Graph domain over five seeds, each an independently drawn graph; solid is baseline,
dashed $+\xtok$; harness budget $K=100$. (a) Single-shot validity. (b) Harness validity.
(c) Composition of invalid outputs at $n=300$ for a single draw and for the harness output.
(d) Mean draws per instance.} %\siwei{I think we care more about +x vs baseline, not the trend with graph size? Similar for b.} \siwei{What is the budget $K$ here?} \siwei{Can we also show the error composition in Harness case?}}
\label{fig:gain}
\end{figure}

Abort supervision can reduce single-shot validity, especially at smaller model size: at
$n=500$, the two $d=120$ models lose about $11$ percentage points, whereas the loss
for $d=384$ is at most $2.8$ points across the tested sizes.
%The cost is larger where abort precision is lower
%(Figure~\ref{fig:gain}a; details in Appendix~\ref{app:diagnostics}).
With retry, however, the $+\xtok$ model has higher harness validity than the baseline
at every graph size for $d=384$ (Figure~\ref{fig:gain}b), reaching
$98.9$--$99.7\%$ compared with $95.0$--$98.3\%$. This gain requires only a small
number of additional draws: averaged over all instances, the $+\xtok$ harness uses
$1.03$--$1.09$ draws per instance, compared with $1.01$ for the baseline, an
increase of only $0.02$--$0.08$ draws (Figure~\ref{fig:gain}d) in expectation.
%\siwei{Similar question about the conditional expectation. Also, describe how large the incresment is.}

Figure~\ref{fig:gain}c shows how the learned abort signal changes the composition
of invalid outputs. In a single draw, many errors that would otherwise appear as
non-edge or wrong-end outputs are replaced by aborts.
The harness then rejects
these aborting draws and resamples, so many of these errors disappear from the
returned output: the rate of returned outputs with an invalid transition falls from
$2.4\%$ for the baseline to $0.4\%$ for $+\xtok$ at $d=384$, and from $5.4\%$ to
$3.6\%$ at $d=120$.  %Thus, abort supervision converts many non-edge failures into detectable
%aborts, and retry prevents them from appearing in the final output.
%\siwei{If we add the error composition in Harness case, we can say more here.}\shi{added}
The gain persists when we match correct-trajectory exposure by doubling $+\xtok$
training updates (Appendix~\ref{app:budget}).

% Single-shot planning remains difficult on instances
% requiring the composition of observed path segments, across architectures and training conditions
% (Figure~\ref{fig:degree}). 
% \siwei{Either move the figure here, or give more detailed explanations. }

\subsection{Countdown}

In this dataset, three numbers must be combined with $+\,, -\,, \times\,, \div$
to reach a target. Intermediate values and targets are restricted to $1$--$100$.
We enumerate all three-number multisets over $1$--$20$ together with all targets
reachable under these restrictions, giving $16{,}117$ distinct instances, and hold
out a disjoint tenth. A valid solution writes one equation per operation,
\texttt{a op b = c}: the two operands are removed from the available numbers and
the result is inserted. Every operation must be correct, and the final result must
equal the target.

To construct an error trajectory, we corrupt the result after one \texttt{=}.
For each trajectory, we choose one of two corruption types with equal probability:
either adding or subtracting a small offset of magnitude $1$--$3$, or replacing the
result with a random value. %We resample until the corrupted value is in range and
%incorrect.
%\siwei{why there are two corruption methods?}
Random replacements often produce obviously implausible results, whereas small
offsets produce plausible-looking arithmetic errors that require checking the
operation more carefully. Mixing the two reduces the chance that the model learns
a shortcut based only on unusually large or small results. The corrupted equation
is the invalid transition and is followed immediately by \xtok; only this final
\xtok\ receives loss on the error trajectory.
%\siwei{needs to describe where the X is, and also how to count the loss.}\shi{added}
We denote by $\rho$ the ratio of error trajectories to correct trajectories.

% The abort precision remains between $89\%$ and $91\%$ throughout.
% \siwei{I think this sentence is better before the description of Figure 3. But not very sure where it should be.}

\begin{figure}[!htbp]
\centering
\includegraphics[width=0.9\textwidth]{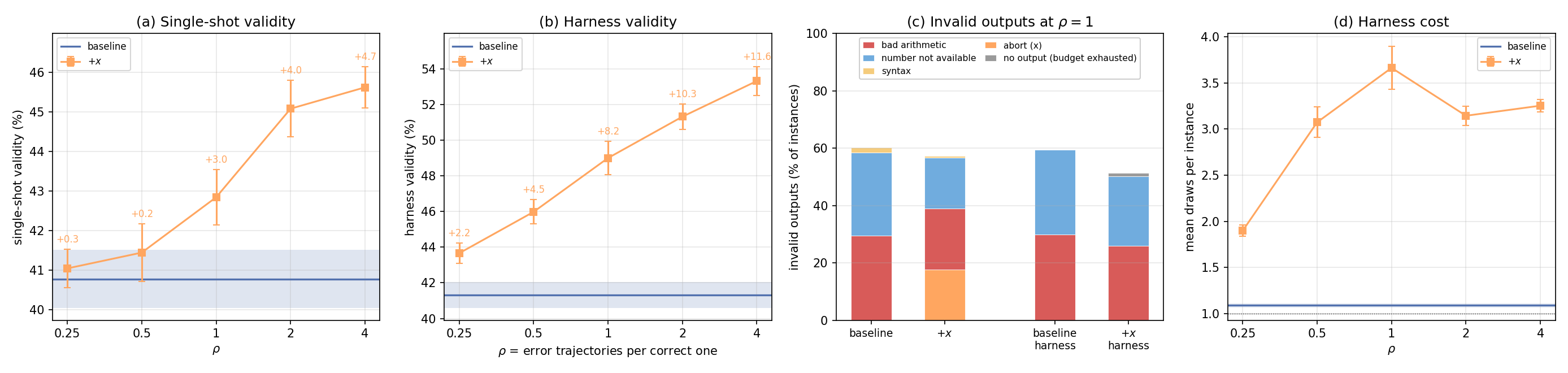}
\caption{Countdown, $1/2/384$, five seeds per $\rho$. (a) Single-shot validity against $\rho$,
with the paired gain over the baseline. (b) Harness validity. (c) Composition of invalid outputs at
$\rho=1$ for a single draw and for the harness output. (d) Mean draws per instance.
In (a), (b) and (d) the baseline is a level band (mean $\pm$ s.e.\ over seeds) because its
training data do not depend on $\rho$.}
\label{fig:countdown}
\end{figure}

At $\rho=1$, %the ratio also used in the graph and maze experiments,
the $+\xtok$ model improves harness validity by
$\mathbf{+8.2 \pm 0.9}$ percentage points (Figure~\ref{fig:countdown}).
This comes at a higher sampling cost: averaged over all instances, the $+\xtok$
harness uses $3.7$ draws per instance, counting an exhausted budget of $100$ in
full, compared with $1.1$ for the baseline. % ($2.5$ draws among accepted instances).
%The remaining errors are mostly ones that the acceptance test cannot detect.
The $+\xtok$ harness fails on $51\%$ of instances: it returns an output with incorrect
arithmetic on $26\%$ and one using an unavailable number on $24\%$, and it exhausts its
budget without accepting an output on $1.2\%$. The baseline harness fails on $59\%$ of
instances (Figure~\ref{fig:countdown}c).

We next vary the amount of abort supervision. Averaging repeated baseline runs
within each seed gives a harness validity of $41.3 \pm 0.7\%$ over five seeds.
The gain from $+\xtok$ increases monotonically from $+2.2 \pm 0.6$ percentage
points at $\rho=0.25$ to $+11.6 \pm 1.3$ points at $\rho=4$
(Figure~\ref{fig:countdown}b). Abort supervision also improves single-shot
validity, by $3.0 \pm 0.5$ percentage points at $\rho=1$ and
$4.7 \pm 0.8$ points at $\rho=4$ (Figure~\ref{fig:countdown}a).
%\siwei{Require a figure. And I am not sure what the following mean.}
This differs from the graph setting, where most of the harness gain comes from
retrying after an abort. On Countdown, validity improves even before retry, suggesting
that abort supervision may also improve representations used to generate valid
equations.

\subsection{Mazes}

In a $10\times10$ maze with random walls, the model starts facing East and emits
relative-direction tokens: \texttt{F} moves forward, \texttt{L} and \texttt{R}
turn left and right, and \texttt{T} turns around. Each token changes direction as
needed and then advances one cell. For each of four independently generated mazes,
we split reachable source--target pairs between training and test sets with
probability $0.5$ each. Training uses correct paths and an equal number of error
trajectories.

A valid output reaches the target without crossing a wall. %, where crossing the maze
%boundary is treated as crossing a wall.
%\siwei{are we going to call both crossing a wall?}\shi{Yes}
Error trajectories contain one wall-crossing move followed by the supervised
abort token.
For mazes, the acceptance test $\mathcal A$ accepts every non-aborting output.
%\siwei{Do you mean the algorithm A?}\shi{Yes}
Thus, the harness retries only when the model emits $\xtok$; wall crossings are
checked only offline when computing validity.

\begin{figure}[!htbp]
\centering
\includegraphics[width=0.70\textwidth]{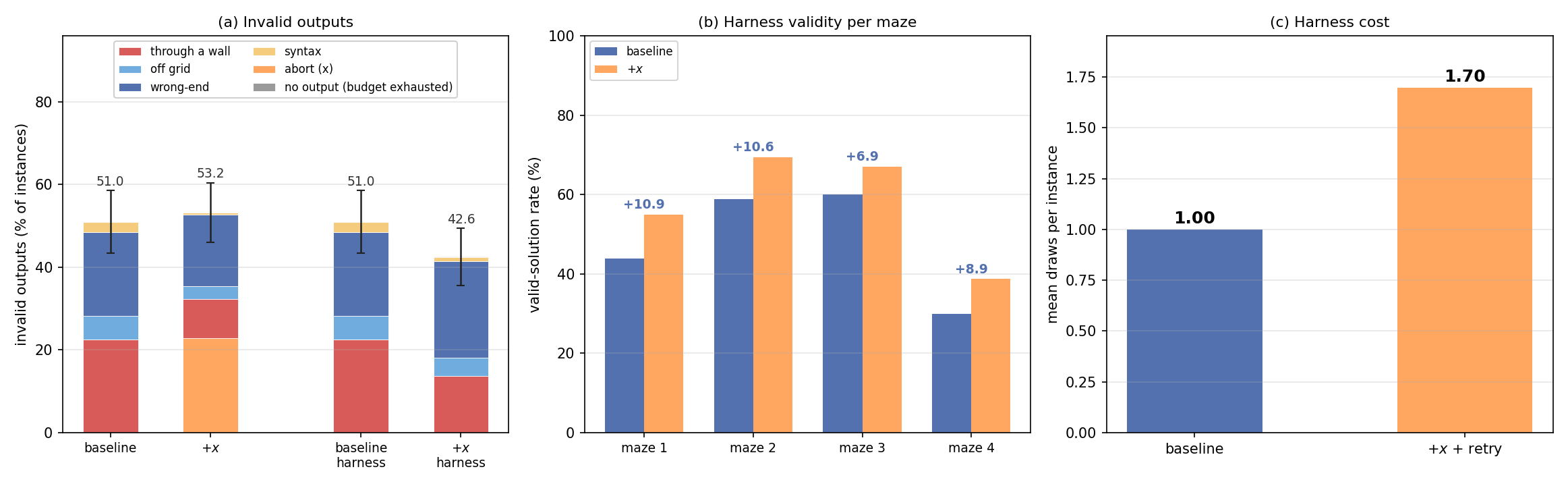}
\caption{Maze task, $1/2/384$, four independently generated mazes. (a) Composition of invalid outputs for a
single draw and the harness output. (b) Harness validity per maze, averaged over two
runs. (c) Mean draws per instance.}
\label{fig:grid}
\end{figure}

Rejecting outputs that contain \xtok\ and drawing again improves harness validity by
$\mathbf{+9.3 \pm 0.9}$ percentage points (Figure~\ref{fig:grid}). This requires
$1.70$ draws per instance on average for the $+\xtok$ harness, %($1.55$ among accepted
%instances)
while the baseline always accepts its first draw.
In single-shot generation, abort supervision reduces the wall-crossing rate from
$22.5\%$ to $9.4\%$, with some of these failures replaced by aborts. The harness
retries aborting draws, reducing wall crossings in returned outputs from
$22.5\%$ for the baseline to $13.7\%$ for $+\xtok$. Wrong-end errors remain at
$20$--$23\%$: neither the learned abort signal nor the acceptance test detects
them (Figure~\ref{fig:grid}a).
%\siwei{One question, why the baseline not the same as baseline harness? Is it because randomness?}\shi{Yes, because of randomness. Now I change the source to the same baseline data.}
%\siwei{Also, after reading the experiments, I feel out we need to mention the difference between abort and best-of-K, i.e., best-of-K require an oracle check to look for a correct answer in K candidates, while ours do not require such an oracle. I haven't read the intro, so if it is contained already, please ignore that. Otherwise we may need a few discussion.}

\section{Conclusion}

Error trajectories can teach a Transformer to detect invalid transitions by
supervising only a final abort token. Our theoretical analysis characterises the
required previous--current interaction and gives a one-layer construction in
which attention retrieves information and the FFN checks transition validity.
The width lower bound further characterises the capacity required within the
specified readout class.
In trained models, abort supervision increases previous-node attention, and the
FFN produces most of the validity-dependent abort signal. Using this learned
signal to reject and retry improves planning reliability on graphs, Countdown
and mazes.

\clearpage

\subsection*{AI use statement}

Generative artificial intelligence (AI) tools were used to assist with parts of the research and writing process. Claude (Anthropic) was used for code debugging and as an aid in literature search, while Claude and Codex (OpenAI) were used for limited drafting and editing of the manuscript. The research questions, problem formulation, theorem statements, proof strategies, experimental design, and interpretation of results were developed by the authors. The authors reviewed and verified all AI-assisted content and take responsibility for the final manuscript and accompanying artifacts.

\subsection*{Reproducibility statement}

Reproducibility details are provided in the paper and appendices.
Section~\ref{sec:setting} describes the architectures, training objective and graph
data, and Section~\ref{sec:endtoend} describes the Countdown and maze tasks and
the retry procedure. Appendix~\ref{app:setting} gives the training and evaluation
protocols, Appendix~\ref{app:diagnostics} details the mechanistic analyses, and
Appendices~\ref{app:controls} and~\ref{app:additional} report additional experiments,
controls and method comparisons. Appendix~\ref{app:theory} provides the explicit
Transformer construction and proofs. We will release the code publicly.

\bibliographystyle{iclr2027_conference}
\bibliography{main}
\clearpage
\appendix
\section*{Appendix}

\section{Theoretical Analysis}
\label{app:theory}

Appendix~\ref{app:illegal} proves Theorem~\ref{thm:illegal} and Appendix~\ref{app:width}
proves Theorem~\ref{thm:width}. We also checked auxiliary steps in Lean~4 using the
Mathlib library: the ReLU identity \eqref{eq:gate} and its sum over
nodes, the bounds on the errors caused by finite attention scores, the bounds on softmax
probabilities, and the counting inequality used in Theorem~\ref{thm:width}. These checks
cover individual lemmas rather than the complete Transformer construction; the
attention-score calculation in \S\ref{app:illegal} and the cited VC-dimension bound of
\citet{vcrelu} are not formalised. A numerical check builds the
Transformer of Appendix~\ref{app:illegal} from its weights at the constants of the proof
and verifies both clauses, the dead-end case and the first step on random DAGs.

\subsection{Construction and Proof of Theorem~\ref{thm:illegal}}
\label{app:illegal}

We use the architecture of \S\ref{sec:architecture} and the notation of \S\ref{sec:theory}.
The construction below has embedding size $d=4|V|+8$ and feed-forward width
$\dff=2|V|$, of which $|V|$ units form the detector.

\begin{thmrestate}
Given a DAG $G$ with adjacency matrix $\Atrue$ and reachability matrix $\Rtrue$, for
every $\varepsilon > 0$ there is a $1$-layer, $2$-head Transformer with embedding size
$O(|V|)$ and feed-forward width $O(|V|)$ that, for every input sequence
``$s\ t\ s\ u_1 \dots u_k$'' for which $(u_0,\dots,u_{k-1})$ is a valid path prefix,
where $u_0=s$, $k \geq 1$, and $u_k \neq t$, generates with probability at least
$1-\varepsilon$ (i) the token \xtok, if $\Atrue_{(u_{k-1},\,u_k)} = 0$, and (ii) otherwise a
node $u_{k+1}$ that is both an out-neighbour of $u_k$ and able to reach $t$, whenever such
a node exists.
\end{thmrestate}

\begin{proof}
Throughout the proofs, $e_v$ denotes the standard basis vector
for coordinate $v$, with dimension determined by the containing space.
Layer normalisations are omitted, as in \citet{alpine}. Write $p = u_{k-1}$, with $u_0 = s$, for
the token preceding $u_k$, so clause~(i) asks the model to detect $\Atrue_{(p,u_k)} = 0$.
We first consider exact attention, with each head assigning unit weight to its
intended position, then bound the effect of finite attention scores.

\paragraph{Embedding.}
Take $M = |V| + 2$ tokens (the nodes, ``\texttt{\symbol{92}n}'', \xtok), per-head size
$d_k = 2|V| + 4$ and $d = 2 d_k = 4|V| + 8$. Coordinates $1,\dots,M$ carry the one-hot
token ($\mathcal{V}$ denotes the node coordinates), coordinate $M+1$ a flag set to a fixed $c_0>0$
on row $2$ only, coordinates $M+2$ and $M+3$ the position $\nu$ and $\nu^2$, and two disjoint
blocks $\mathcal{R}, \mathcal{P}$ of $|V|$ coordinates in the second half hold the two
head outputs. All other embedding coordinates are zero. Positions are numbered
starting at $1$, so the target occupies position $2$.
Thus row $\nu$ of $H_0$ is $e_{\tau_\nu}$ on the token coordinates, with
$\tau_\nu$ the token at position $\nu$, plus $(c_0\mathbf{1}[\nu=2],\, \nu,\, \nu^2)$. Writing
$\Pi = \sum_{v \leq M} e_v$, every row has $\langle (H_0)_{\nu,:}, \Pi\rangle = 1$, which
is how queries independent of $\nu$ are formed. The DAG path $(u_0,\dots,u_{k-1})$
has distinct nodes. The final node $u_k$ need not extend that path, so
$k\leq |V|$ and the length $N=k+3$ satisfies $N\leq N_{\max}:=|V|+3$. All constants
below depend only on $N_{\max}$ and $\varepsilon$.

\paragraph{Heads.}
For $\gamma>0$, Head~1 has query
$\mathbf{1}_{d_k}$ at every row (from $\Pi$) and key
$\gamma c_0\mathbf{1}[m=2]\mathbf{1}_{d_k}$ at row $m$, so after scaling its score at
column $m$ is $\beta_1\mathbf{1}[m=2]$ with $\beta_1 := \sqrt{d_k}\,\gamma c_0$. Its
value projection is $\Rtrue$ from the node coordinates into $\mathcal{R}$, so at exact
attention it writes $\Rtrue_{(t,:)}$ into $\mathcal{R}$. For $\lambda>0$, Head~2 uses
key $\lambda\sqrt{d_k}(m^2,m)$ at row $m$ and query $(-1,2(\nu-1))$ at row $\nu$,
with all remaining query and key coordinates zero. These are linear projections
of the embeddings. After the standard $1/\sqrt{d_k}$ scaling, the score is
$\lambda(-m^2+2m(\nu-1))=\lambda\big((\nu-1)^2-(m-(\nu-1))^2\big)$.
At the final position $\nu=N$, the first term is constant in $m$, so under the causal
mask the score is maximised at $m=N-1$ and every other column is lower by at least
$\lambda$. Its value projection is $\Atrue$ from the node coordinates into
$\mathcal{P}$, so at exact attention it writes the row $\Atrue_{(p,:)}$ into $\mathcal{P}$. $\WA$ places the
two outputs in $\mathcal{R}$ and $\mathcal{P}$ and nothing in $\mathcal{V}$. Both
value projections are zero on non-node token coordinates and on positional
coordinates; all unspecified projection entries are zero. Thus, after
the residual connection $\tilde H_{N,\mathcal{V}} = e_{u_k}$ exactly.

Both heads have one column leading all others by a margin, $\beta_1$ and $\lambda$
respectively, among at most $N_{\max}$ columns; the softmax weight off that column is
therefore at most $(N_{\max}-1)e^{-\beta_1}$ and $(N_{\max}-1)e^{-\lambda}$
by the softmax bound used below.

\paragraph{Feed-forward network.}
Write $i=u_k$ and $h=\tilde H_{N,:}$. At exact attention, the three blocks read by
the FFN are
\[
h_{\mathcal R}\oplus h_{\mathcal P}\oplus h_{\mathcal V}
=\Rtrue_{(t,:)}\oplus\Atrue_{(p,:)}\oplus e_i,
\]
where $\oplus$ denotes concatenation. The first linear map $W_1$ extracts the current
node's adjacency row from $e_i$: its $v$th entry is
$\sum_{j\in V}e_i(j)\Atrue_{(j,v)}=\Atrue_{(i,v)}$.
We give $W_1$ two groups of $|V|$ columns. The entries of $b_1$ are $0$ on the first
group and $-1$ on the second, producing two groups of inputs to the ReLU activation:
\[
\begin{aligned}
z^{\mathrm{plan}}_v
&=h_{\mathcal R,v}+\sum_{j\in V}h_{\mathcal V,j}\Atrue_{(j,v)}
 =\Rtrue_{(t,v)}+\Atrue_{(i,v)},\\
z^{\mathrm{det}}_v
&=h_{\mathcal V,v}+h_{\mathcal P,v}-1
 =e_i(v)+\Atrue_{(p,v)}-1.
\end{aligned}
\]
All entries of $W_1$ not specified by these expressions are zero. Applying ReLU to
both groups in parallel gives
\[
\begin{aligned}
q_v&=\mathrm{ReLU}(z^{\mathrm{plan}}_v)
    =\Rtrue_{(t,v)}+\Atrue_{(i,v)}\in\{0,1,2\},\\
d_v&=\mathrm{ReLU}(z^{\mathrm{det}}_v)
    =\mathbf{1}[i=v]\Atrue_{(p,v)}.
\end{aligned}
\]
Thus $q_v=2$ exactly for the out-neighbours of $i$ that can reach $t$, $q_v=1$ for
nodes with one of the two properties, and $q_v=0$ otherwise, as in the
adjacency--reachability account of \citet{alpine}; the planning inputs are
non-negative, so their ReLU is the identity.
The detection units implement \eqref{eq:gate}: only $d_i$ can be nonzero,
and $\sum_v d_v=\Atrue_{(p,i)}$.
The current node's adjacency row is thus read linearly, and the same ReLU layer
implements the gate.

Choose $c\geq4$. In $W_2$, the output row for $q_v$ is $c e_v$, while the row for
$d_v$ is $-6c e_{\xtok}$; set $b_2=3c e_{\xtok}$. The planning units write only to
node coordinates, and the detection units write only to the abort coordinate. The
abort logit is $3c-6c\sum_v d_v$, hence $3c$ on a non-edge and $-3c$ on an edge,
independent of reachability. The feed-forward width is $\dff=2|V|$.

\paragraph{Output at exact attention.}
Choose $\WU$ to read the token coordinates. The vocabulary
logits at row $N$ of $\mathrm{FFN}(\tilde H) + \tilde H$ are then
\begin{equation}
\begin{aligned}
z_v &= c\,\bigl(\Rtrue_{(t,v)}+\Atrue_{(u_k,v)}\bigr) + \mathbf{1}[v=u_k]
&& (v \in V),\\
z_{\texttt{\symbol{92}n}} &= 0, \\
z_{\xtok} &= 3c - 6c\,\Atrue_{(p,u_k)} .
\end{aligned}
\label{eq:logits2}
\end{equation}
A DAG has $\Atrue_{(u_k,u_k)} = 0$, so a node logit is $2c$ exactly when the node is an
out-neighbour of $u_k$ that reaches $t$, at most $c+1$ when it has one of the two
properties, and at most $1$ otherwise. Writing $B$ for the set of desired output tokens,
in clause~(i) $B = \{\xtok\}$ at $3c$ against at most $2c$ for every other token. In
clause~(ii), $B$ is the nonempty set of out-neighbours that can reach $t$, each with
logit $2c$, against at most $c+1$ for other nodes, $0$ for \texttt{\symbol{92}n}, and
$-3c$ for \xtok. In both cases
\begin{equation}
\min_{v\in B} z_v - \max_{v\notin B} z_v \;\geq\; g := c/2
\label{eq:gap}
\end{equation}
because the respective gaps are at least $c$ and $c-1$. If the most recent transition
is valid but no out-neighbour can reach $t$, the abort logit is still $-3c$, below every
non-abort logit. Taking $B$ to be all non-abort tokens then gives a gap of at least
$3c$, so \eqref{eq:gap} also holds in this case. Thus even at a dead end, a valid
preceding transition suppresses abort; the next-node guarantee applies only when an
out-neighbour that can reach $t$ exists, as stated in clause~(ii). At such a dead end
the nodes with $q_v=1$, namely the out-neighbours of $u_k$ and the nodes that can reach
$t$ (a set that contains $t$), have logit at least $c$, and every other node has logit at
most $1$. Taking $B$ to be this set gives a gap of at least $c-1\geq g$, so the model
samples from this narrower set.

\paragraph{Finite attention scores.}
Two FFN groups must be considered. The planning group reads $\mathcal{R}$, a convex combination of
rows of $\Rtrue$ that puts weight at most $(N_{\max}-1)e^{-\beta_1}$ off the row $t$.
The adjacency input is exact, and ReLU cannot increase the difference between two
scalar inputs, so the node logits change by at most
$c(N_{\max}-1)e^{-\beta_1}$. This group has no effect on the abort logit.
The detection group reads $\mathcal{V}$ exactly and $\mathcal{P}$ as a
convex combination of rows of $\Atrue$, so unit $v$ sees some $\tilde a_v \in [0,1]$ in
place of $\Atrue_{(p,v)}$, with $|\tilde a_v - \Atrue_{(p,v)}| \leq (N_{\max}-1)e^{-\lambda}$.
For $v \neq u_k$ the argument $\tilde a_v - 1$ is non-positive and the unit stays at $0$;
for $v = u_k$ the unit equals $\tilde a_{u_k}$. Through its output
row this moves only the \xtok\ logit, by at most $6c(N_{\max}-1)e^{-\lambda}$.
Together, these deviations bound every logit
change by
\[
\eta := c(N_{\max}-1)e^{-\beta_1} + 6c(N_{\max}-1)e^{-\lambda}.
\]
Taking
\[
\beta_1 \geq \log(16 N_{\max}), \qquad \lambda \geq \log(96 N_{\max}),
\]
gives $\eta \leq c/16 + c/16 = g/4$, so every logit is within $g/4$ of
\eqref{eq:logits2} and the gap is still at least $g - 2\eta \geq g/2$.
The softmax probability assigned outside the desired set
$B$ is at most $(M-1)e^{-g/2} = (M-1)e^{-c/4}$, which is at most $\varepsilon$ once
$c \geq 4\log((M-1)/\varepsilon)$. So $c = \max\{4,\, 4\log((M-1)/\varepsilon)\}$ and
the two attention-score bounds above give the theorem, with $d = 4|V|+8$ and $\dff = 2|V|$.
The same bound with $B$ equal to all non-abort tokens shows
$\Pr(\xtok)\leq\varepsilon$ after a valid transition into a dead end. At finite
attention the abort logit is exactly $3c-6c\tilde a_{u_k}$ and still has no
reachability input. Its softmax probability shares a denominator with the node logits,
but whether abort is the highest-logit token is determined by transition validity
under the bounds above.

\paragraph{Boundary cases.}
The theorem concerns positions after at least one generated transition and before
the current node equals the target. It does not specify the initial prediction
of $s$ from ``$s\ t$'' or the end-of-line prediction after reaching $t$.
At the first step, $k=0$ and the input is ``$s\ t\ s$''. Head~2 then attends to
position~$2$, so the detector reads the pair $(t,s)$, which is not a transition of the
path, and $\Atrue_{(t,s)}=0$ because $s\rightsquigarrow t$ in a DAG; as constructed, the
model would emit \xtok. One further value map covers this step without changing $d$ or
$\dff$: head~2 also copies the flag coordinate into the abort coordinate with weight
$-6c/c_0$. At $\nu=3$, head~2 places weight at least $1-(N_{\max}-1)e^{-\lambda}$ on
position~$2$, so the abort logit is at most $-3c+6c(N_{\max}-1)e^{-\lambda}$, while the
planning units select an out-neighbour of $s$ that reaches $t$, as in clause~(ii). At
later positions the map lowers the abort logit by at most $6c(N_{\max}-1)e^{-\lambda}$;
taking $\lambda\geq\log(192N_{\max})$ in place of $\log(96N_{\max})$ keeps the total
change of the abort logit at most $c/16$, so the bounds above are unchanged.

The case analysis uses only that the reachability matrix has entries in $\{0,1\}$.
It therefore also holds with $\Robs$ in place of $\Rtrue$, the setting of
Remark~\ref{rem:observed-retry}. Detection is unchanged. After a valid transition, or at
the first step, if no out-neighbour of the current node is observed to reach $t$, the
model outputs with probability at least $1-\varepsilon$ an out-neighbour of the current
node or a node observed to reach $t$, provided this set is nonempty.
\end{proof}

\subsection{Lower Bound on FFN Width}
\label{app:width}

We prove Theorem~\ref{thm:width} in the setting of \S\ref{sec.widthlowerbound}: the FFN
input is $x_{p,i,t}=e_p\oplus e_i\oplus e_t\in\mathbb{R}^{3n}$ with $n=|V|$, and the FFN's
contribution to the abort logit is \eqref{eq:FFN}.

\paragraph{Scalar form and fixed target.}
With $W_1^{(p)},W_1^{(i)},W_1^{(t)}$ the row blocks of $W_1$ from
\S\ref{sec.widthlowerbound}, let $w_h=\bigl(W_2\,\WU[:,\xtok]\bigr)[h]$ be the weight with
which unit $h$ enters the abort logit and $\beta=b_2\,\WU[:,\xtok]$. Because each block of
$x_{p,i,t}$ is one-hot, \eqref{eq:FFN} is the scalar ReLU network
\begin{equation}
z_{\xtok}^{\mathrm{FFN}}(p,i,t)=\beta+\sum_{h=1}^{r} w_h\operatorname{ReLU}\bigl(W_1^{(p)}[p,h]+W_1^{(i)}[i,h]+W_1^{(t)}[t,h]+b_1[h]\bigr).
\label{eq:scalar-ffn}
\end{equation}
We take $z_{\xtok}^{\mathrm{FFN}}>0$ to mean ``invalid''; the argument is identical under
the opposite convention. Fix a target $t_0$ and absorb $W_1^{(t)}[t_0,h]$ into the bias,
$b'_h=b_1[h]+W_1^{(t)}[t_0,h]$, giving
\begin{equation}
F_{t_0}(p,i):=z_{\xtok}^{\mathrm{FFN}}(p,i,t_0)=\beta+\sum_{h=1}^{r} w_h\operatorname{ReLU}\bigl(W_1^{(p)}[p,h]+W_1^{(i)}[i,h]+b'_h\bigr).
\label{eq:fixed-target-ffn}
\end{equation}
Let $\mathcal{G}_r$ be the class of Boolean functions $\mathbf{1}[F_{t_0}(p,i)>0]$ on
$V\times V$, with $W_1^{(p)},W_1^{(i)}\in\mathbb{R}^{n\times r}$, $b',w\in\mathbb{R}^r$ and
$\beta\in\mathbb{R}$ arbitrary.
An FFN that is correct for every target must also be correct for this fixed target.
Its decision rule belongs to $\mathcal{G}_r$, so a lower bound for this class also
applies when the FFN receives the target as an input.

\begin{thmrestatew}
Suppose that a width-$r$ one-hidden-layer ReLU FFN receives the previous node
$p$, current node $i$, and target $t$ as the input $e_p\oplus e_i\oplus e_t$.
If, for every DAG on $|V|$ nodes, there exist FFN parameters for which the sign
of $z_{\xtok}^{\mathrm{FFN}}(p,i,t)$ correctly determines whether $(p,i)\in E$
for every $p,i,t\in V$, then $r=\Omega\bigl(|V|/\log^2|V|\bigr)$.
\end{thmrestatew}

\begin{proof}
Fix $t_0$ and let $\mathcal{G}_r$ be the class defined above. The proof
compares how many labellings of $V\times V$ the class $\mathcal{G}_r$ can
induce with how many it must induce.

\emph{Step 1: number of parameters.}
A member of $\mathcal{G}_r$ is specified by the $2nr$ entries of $W_1^{(p)}$ and
$W_1^{(i)}$, the $r$ entries each of $b'$ and $w$, and $\beta$. Hence
\[
P
=
r(2n+2)+1
\le
4rn
\qquad(n\ge2,\ r\ge1),
\]
since $r(2n+2)+1\le 4rn$ is equivalent to $1\le r(2n-2)$.

\emph{Step 2: VC dimension.}
$\mathcal{G}_r$ contains the binary decisions obtained by thresholding a
one-hidden-layer ReLU network with $P$ real parameters, evaluated on the finite input set
$\{e_p\oplus e_i\}\subset\mathbb{R}^{2n}$. The Vapnik--Chervonenkis (VC) dimension
is the largest number of inputs on which a function class can realise every
binary labelling. The VC dimension of the sign of a
piecewise-linear network with $P$ parameters and a constant number of layers is
at most $C\,P\log P$ for an absolute constant $C$~\citep{vcrelu}; restricting
the domain cannot increase it. By Step~1 and monotonicity of
$P\log P$,
\begin{equation}
D:=\operatorname{VCdim}(\mathcal{G}_r)\le C\,P\log P\le 4C\,rn\log(4rn).
\label{eq:vc-upper}
\end{equation}

\emph{Step 3: how many labellings $\mathcal{G}_r$ can induce.}
The domain $V\times V$ has $m=n^2$ points. By the Sauer--Shelah lemma, a class
of VC dimension $D$ induces at most $\sum_{k=0}^{D}\binom{m}{k}\le(m+1)^D$
distinct labellings of an $m$-point set. Thus the number induced by
$\mathcal{G}_r$ on $V\times V$ satisfies
\begin{equation}
\bigl|\mathcal{G}_r\bigr|
\le
(n^2+1)^D.
\label{eq:number-upper}
\end{equation}

\emph{Step 4: how many labellings it must induce.}
Fix the node ordering $1,\dots,n$ and consider the graphs whose edges all go
from a smaller index to a larger index. Every subset of the $\binom{n}{2}$
pairs $\{(p,i):p<i\}$ is such a graph, and each is a DAG because its edges
follow a fixed topological order. Distinct subsets have distinct adjacency
matrices, so there are $2^{\binom{n}{2}}=2^{n(n-1)/2}$ distinct adjacency
relations. By hypothesis each of these DAGs is detected by some FFN, and by
the fixed-target reduction the restriction of that FFN's decision rule to
$t=t_0$ is a member of $\mathcal{G}_r$ whose labelling of $V\times V$ is the
indicator of the non-edges of that DAG. Distinct DAGs give distinct
labellings, so
\begin{equation}
\bigl|\mathcal{G}_r\bigr|
\ge
2^{n(n-1)/2}.
\label{eq:number-lower}
\end{equation}

\emph{Conclusion.}
Combining \eqref{eq:number-upper} and \eqref{eq:number-lower} and taking
base-two logarithms,
\[
D
\ge
\frac{n(n-1)}{2\log_2(n^2+1)}
=
\Omega\!\left(\frac{n^2}{\log n}\right).
\]
With \eqref{eq:vc-upper} this gives $rn\log(4rn)=\Omega(n^2/\log n)$, that is,
\[
r\log(4rn)=\Omega\!\left(\frac{n}{\log n}\right).
\]
If $r>n$, the claimed asymptotic lower bound already holds. Otherwise
$r\le n$, so $\log(4rn)\le\log(4n^2)=O(\log n)$, and dividing the display by
$\log(4rn)$ gives $r=\Omega(n/\log^2 n)$. In either case
$r=\Omega\bigl(|V|/\log^2|V|\bigr)$.
\end{proof}

\paragraph{Matching upper bound.}
The detector of Theorem~\ref{thm:illegal} lies in this class with $r=|V|$. Setting
$W_1^{(p)}[p,v]=\Atrue_{(p,v)}$, $W_1^{(i)}[i,v]=\mathbf{1}[i=v]$, $W_1^{(t)}=0$,
$b_1[v]=-1$, $w_v=-6c$ and $\beta=3c$, unit $v$ computes \eqref{eq:gate} and
$z_{\xtok}^{\mathrm{FFN}}(p,i,t)=3c-6c\,\Atrue_{(p,i)}$, which is positive exactly on
non-edges. In the Transformer of Appendix~\ref{app:illegal} the map
$e_p\mapsto\Atrue_{(p,:)}$ is supplied by head~2; here it is absorbed into $W_1^{(p)}$.
Width $|V|$ therefore suffices for every DAG, and Theorem~\ref{thm:width} shows that
this is optimal up to a factor $O(\log^2|V|)$.

\section{Experimental Details and Extended Results}
\label{app:experiments}

\subsection{Setup and Evaluation Details}
\label{app:setting}

\paragraph{Training objective.}
The trained models apply layer normalisation before attention, before the FFN and before
unembedding. Let $\mathcal D$ be the training set. For each sequence $u=(u_1,\dots,u_N)$,
the supervision mask $\mu\in\{0,1\}^N$ marks which tokens receive a loss:
\begin{equation}
\mathcal{L}(\mathcal{D}) \;=\; -\sum_{u \in \mathcal{D}} \sum_{q=1}^{N-1} \mu_{q+1}\,
\log \Pr\!\left[u_{q+1} \mid u_{1:q}\right].
\label{eq:loss}
\end{equation}
Correct trajectories supervise every non-padding next token after the first; an error
trajectory $(s,\ t,\ s,\ u_1,\dots,u_k,\ w,\ \xtok)$, with $u_0=s$ when $k=0$, supervises
only the final \xtok. Minibatch losses are averaged over supervised labels.

\paragraph{Graph data.}
\label{par:fairness}
The distinct non-edges used in the error trajectories cover $82$--$93\%$ of all non-edges,
more at larger $n$. The $+\xtok$ data sets range from $67{,}000$ sequences ($2.1$ million
padded token slots) at $n=100$ to $2.49$ million sequences ($79.6$ million slots) at
$n=500$. On the $n=300$ data, $37.4\pm0.3\%$ of the positions supervised to emit \xtok\
also end at a node from which the target is unreachable, so the abort label is not a
dead-end label. A correct trajectory carries $6.1$ supervised labels on average at
$n=100$ and $8.1$ at $n=500$, whereas an error trajectory carries one; with half of each
$+\xtok$ batch made of error trajectories, the $+\xtok$ model sees $0.50\times$ the
baseline's correct trajectories per update in expectation (measured at $n=100$, $300$ and $500$).
This is a ratio of trajectory counts, not a ratio of loss weights.
Appendix~\ref{app:budget} doubles $+\xtok$ updates to match correct-trajectory exposure.

\paragraph{Output classification.}
Prompted with \texttt{s t}, the model emits tokens until end-of-line or, for a $+\xtok$
model, the abort token, subject to the generation-length limit. An output is
\textsf{valid} if it is a path from $s$ to $t$ containing only graph edges, and
\textsf{invalid} otherwise. Invalid outputs are classified in the evaluator's order as
\textsf{abort} (it contains \xtok), \textsf{syntax} (too few node tokens),
\textsf{wrong-end} ($v_0\ne s$ or $v_L\ne t$ for the output $v_0,\dots,v_L$) or
\textsf{non-edge} (correct endpoints but some $(v_{q-1},v_q)\notin E$). Every output
containing \xtok\ counts as \textsf{abort}, whether or not the abort is justified,
including one whose last transition before \xtok\ is a non-edge; a wrong-end output may
also contain a non-edge.

\paragraph{Evaluation, abort precision and selective accuracy.}
\label{sec:setup}
Harness results use $3{,}000$ instances drawn with replacement from the held-out set, with
$K=100$, in all three domains of \S\ref{sec:endtoend}. Graph single-shot validity is
evaluated on every held-out instance, the Countdown single-shot results use the same
$3{,}000$ instances as the harness, and the maze results average two independent sampling
runs per maze. An abort emitted after a prefix $y_{1:k}$ is
\begin{equation}
\textsf{justified} \iff \text{no completion of } y_{1:k}
\text{ is a valid output for the instance},
\label{eq:justified}
\end{equation}
and abort precision is the fraction of emitted aborts that are justified. In the
graph domain a prefix admits a valid completion exactly when it starts at $s$, follows
edges and ends at a node that can reach $t$, which we check exhaustively on $G$. For the
fitted models of Appendix~\ref{app:additivefit}, we instead report the fraction of
total abort probability assigned to prefixes that admit a valid completion.

We also compare aborting with rejecting outputs that have low mean token
log-probability. Coverage is the fraction of instances whose outputs are retained;
selective accuracy is the fraction of those retained outputs that are valid.
For $+\xtok$, we retain every non-aborting output. For the baseline, we discard the
lowest-scoring outputs until the same fraction is retained. These measurements
use one draw per instance, without retry.

\paragraph{Compute.}
Graph models are trained in bfloat16 on one NVIDIA H200 GPU. The two $d{=}120$
architectures have the same $0.2$ million parameters, because increasing the head count
divides the same embedding size among more heads; the $d=384$ model has about $1.9$
million.

\subsection{Additional Mechanistic Results}
\label{app:diagnostics}

\begin{figure}[!htbp]
\centering
\includegraphics[width=0.86\textwidth]{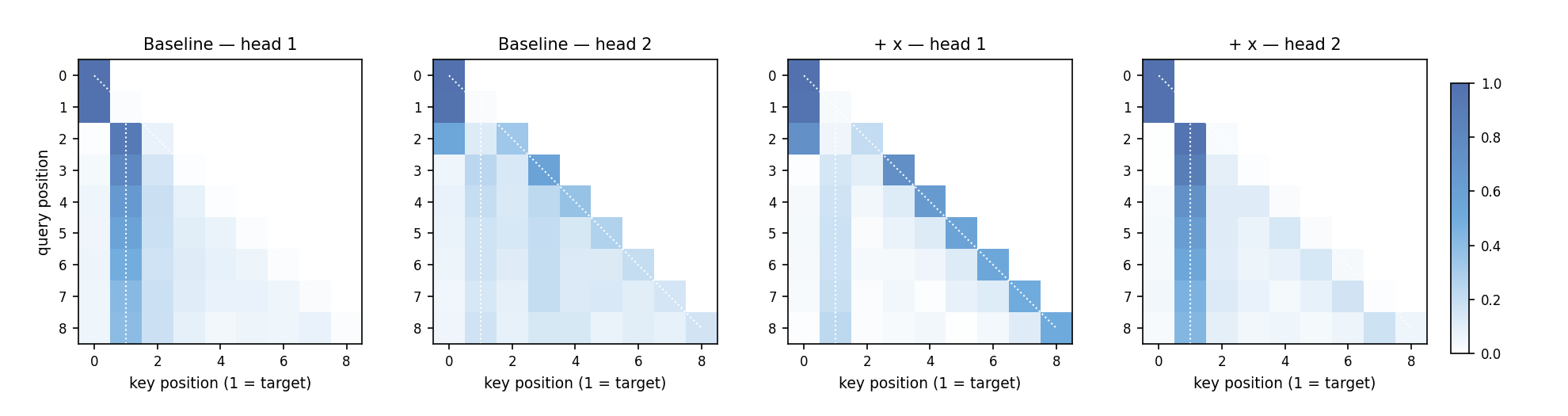}
\caption{Average per-head attention for baseline and $+\xtok$ models, $1/2/384$ at $n=300$.
One representative seed is shown because head identities can swap across runs.}
\label{fig:heads}
\end{figure}

\paragraph{Interventions on previous-node attention and the FFN.}
\label{app:causal}
The interventions of \S\ref{sec:previous-attention} use the $1/2/384$ $+\xtok$ models of
six $n=300$ graphs: the five evaluation seeds and the graph of Figure~\ref{fig:heads}. For
each graph, we take $2{,}000$ held-out valid prefixes for which a previous-node swap
exists, append a non-successor node and measure $P(\xtok)$ at the next position.
Removing attention to a position sets its weight at the final query to zero and
renormalises the remaining weights; the main head is the head with more previous-node
attention on invalid prefixes. Removing the FFN output sets it to zero before the
residual addition. Means $\pm$ s.e.\ are over graphs.

Table~\ref{tab:causal} adds controls and the full sample to Table~\ref{tab:causal-main}.
On target-reachable prefixes, removing target or current-node attention leaves
$P(\xtok)$ at $0.911$ or $0.997$, and removing previous-node attention in the other head
leaves it at $0.861$, so the effect is specific to previous-node attention, mostly in the
main head. On the other $81.7\pm0.3\%$ of prefixes the appended node cannot reach the
target, so an abort remains justified after the swap; the model still aborts there with
mean probability $0.72\pm0.01$, which keeps the all-prefix column near $0.6$. On valid
prefixes, removing previous-node attention in both heads makes $P(\xtok)>0.5$ on only
$2.4\pm0.7\%$ of prefixes.

As a consistency check, we also copy the complete attention output at the final
position between the invalid and swapped prefixes. Because the model has only
one layer, the final token $w$ and its positional embedding are identical in the
two prefixes, and the attention output is the only component at that position
that depends on the earlier tokens before the FFN. The patched prefix therefore
gives the same abort decision as the corresponding swapped input.
\siwei{since there is only one layer? I believe copying the attention output is the same as using another input, is that correct?}\shi{Yes, that is correct.}
On target-reachable prefixes, copying the swapped prefix's attention output into
the invalid prefix gives $P(\xtok)=0.001$, as does the previous-node swap itself
(Table~\ref{tab:causal}). This checks the intervention implementation rather than
providing an independent test of the mechanism.

\begin{table}[!htbp]
\centering
\caption{Interventions after an injected invalid transition, six $n=300$ graphs: mean
$P(\xtok)$ $\pm$ s.e.\ over all prefixes and over those whose target remains reachable.
Removing target or current-node attention, and the valid prefix, are controls.}
\label{tab:causal}
\small
\begin{tabular}{lcc}
\toprule
condition & all prefixes & target reachable \\
\midrule
invalid prefix, no intervention & $0.997 \pm 0.000$ & $0.987 \pm 0.003$ \\
remove previous-node attention, main head & $0.594 \pm 0.037$ & $0.089 \pm 0.086$ \\
remove previous-node attention, other head & $0.949 \pm 0.037$ & $0.861 \pm 0.097$ \\
remove previous-node attention, both heads & $0.604 \pm 0.007$ & $0.008 \pm 0.002$ \\
remove target attention, both heads & $0.729 \pm 0.081$ & $0.911 \pm 0.046$ \\
remove current-node attention, both heads & $0.998 \pm 0.001$ & $0.997 \pm 0.001$ \\
drop the FFN output & $0.000 \pm 0.000$ & $0.000 \pm 0.000$ \\
swap the previous node (last transition valid) & $0.590 \pm 0.009$ & $0.001 \pm 0.000$ \\
patch attention output from the valid swap & $0.590 \pm 0.009$ & $0.001 \pm 0.000$ \\
valid prefix & $0.000 \pm 0.000$ & $0.000 \pm 0.000$ \\
\bottomrule
\end{tabular}
\end{table}

\paragraph{Direct logit attribution.}
\label{app:dla}
At the final position, write the residual stream before the final layer normalisation as
$H = Y + M$, where $Y = X+\operatorname{MHA}(\operatorname{LN}_1(X))$ is the stream entering
the FFN and $M$ is the FFN output, as in \S\ref{sec:main-dla}. With $w_{\xtok}$ the abort
column of $\WU$ and $g$ the scale of the final layer normalisation, which has no bias, the
abort logit is $w_{\xtok}^{\top}\bigl[g\odot(H-\operatorname{mean}(H))/\operatorname{std}(H)\bigr]$,
where $\operatorname{std}(H)$ includes the numerical stabiliser. Centring is linear and
$\operatorname{std}(H)$ is a single scalar for the forward pass, so the logit splits exactly
into $z^{Y}_{\xtok}+z^{M}_{\xtok}$ with
\[
z^{C}_{\xtok}=w_{\xtok}^{\top}\Bigl[g\odot\frac{C-\operatorname{mean}(C)}{\operatorname{std}(H)}\Bigr],
\qquad C\in\{Y,M\}.
\]
For the abort margin, replace $w_{\xtok}$ by $w_{\xtok}-w_{v^*}$, where $v^*$ is the node
with the highest logit on that prefix. The trained FFN has no output bias, so
$M=\sum_h\alpha_h W_2[h,:]$ with $\alpha_h$ the activation of unit $h$, and the same formula
gives each unit's contribution. Table~\ref{tab:dla} reports these quantities on the
prefixes of Table~\ref{tab:causal}.

\begin{table}[!htbp]
\centering
\caption{Direct logit attribution at the abort position, means over six $n=300$ graphs (s.e.\ at
most $0.4$). Contributions of the pre-FFN stream $X+\operatorname{MHA}(\operatorname{LN}_1(X))$
and FFN output $M$ to the abort logit and margin (abort logit minus highest
node logit); ``swap-valid'' swaps the previous node to make the last transition valid.}
\label{tab:dla}
\footnotesize
\begin{tabular}{lccccc}
\toprule
 & & \multicolumn{2}{c}{to abort logit} & \multicolumn{2}{c}{to margin} \\
\cmidrule(lr){3-4}\cmidrule(lr){5-6}
Prefix & Abort logit & pre-FFN & FFN & pre-FFN & FFN \\
\midrule
valid continuation & $3.0$ & $-0.95$ & $+4.0$ & $-1.6$ & $-8.1$ \\
swap-valid, target reachable & $0.3$ & $-1.15$ & $+1.5$ & $-1.7$ & $-10.2$ \\
invalid, target reachable & $29.0$ & $-0.68$ & $+29.6$ & $-1.0$ & $+22.4$ \\
invalid, target unreachable & $28.8$ & $-0.46$ & $+29.2$ & $-0.7$ & $+22.0$ \\
\bottomrule
\end{tabular}
\end{table}

The pre-FFN contribution stays near $-1$ in all four conditions, including the valid
continuation and the target-unreachable invalid prefixes. On valid prefixes the FFN's
contribution to the abort logit is positive on $90\%$ of prefixes, while its margin
contribution is $-8.1$: it raises the highest node logit more than the abort logit. This
polarity is the reverse of the construction in Appendix~\ref{app:illegal}, where a constant
abort bias is cancelled by units that fire after a valid transition. In the trained models
the ten units with the most negative contributions on valid prefixes sum to only
$-1.1\pm0.1$ there, and to $+8.4\pm1.7$ on invalid prefixes. A detection unit
$\mathrm{ReLU}(\mathbf{1}[i=v]-\Atrue_{(p,v)})=\mathbf{1}[i=v](1-\Atrue_{(p,v)})$, which fires
after an invalid transition, implements the same check with the same width when its output
weight is $+6c$ and the abort bias is $-3c$. Most of the
FFN contribution on invalid prefixes comes from a small set of units: $16$ units exceed
$0.25$ when the target remains reachable ($22$ when it does not), the ten largest carry
$63$--$87\%$ of the FFN contribution, and $16$ ($31$) units reach $90\%$ of it.

\paragraph{Additive fits.}
\label{app:additivefit}
Could the supervised abort decision be reproduced by separate additive contributions of
the previous node $p$, current node $i$ and target $t$? We fit, to the supervised training
labels of the $n=300$ graph of Figure~\ref{fig:heads} rather than to the trained model,
the additive-logit model
\begin{equation}
z(\ell) \;=\; W^{M}_{(i,\ell)} + W^{V}_{(t,\ell)} + W^{P}_{(p,\ell)} ,
\label{eq:additive3}
\end{equation}
where $\ell$ is a candidate output token and $W^M,W^V,W^P\in\mathbb{R}^{|V|\times M}$ are
free matrices: a linear readout of three one-hot inputs with no term depending jointly on
$p$ and $i$. This is the form computed by a fixed-attention Transformer in which two heads
copy the target and previous-node representations, the residual supplies the current
node, and the remaining maps are linear. With $W^P=0$ it is the additive model of
\citet{alpine}, which gives every prefix with the same $(i,t)$ the same next-token
distribution. The abort probability of \eqref{eq:additive3} is a softmax over all
candidate logits rather than a thresholded additive score, so
Proposition~\ref{thm:additive} does not apply directly; we test the fit empirically.

The fit uses every supervised position whose prefix ends at $u_0=s$ or at a later node,
$2{,}784{,}440$ positions grouped by $(p,i,t)$: $425{,}002$ follow an invalid transition,
and the rest come from correct trajectories and all admit a valid completion. The logits
are linear in the fitted matrices, so the objective is convex; Adam followed by L-BFGS
reaches $\|\nabla\overline{\mathcal L}\|_\infty\le6.3\times10^{-7}$ for the loss averaged
over these positions. Let $\mathcal{S}$ be the multiset of these prefixes and
$\mathcal{C}\subseteq\mathcal{S}$ those admitting a valid completion (source-only prefixes,
for which the fit sets $p=i=s$, belong to $\mathcal{C}$). The \emph{unjustified-abort mass
fraction} $\sum_{y\in\mathcal C}P(\xtok\mid y)/\sum_{y\in\mathcal S}P(\xtok\mid y)$ is the
share of total abort probability placed on prefixes that could still be completed
correctly; we also report the mean $P(\xtok\mid y)$ over
$\mathcal{I}=\mathcal{S}\setminus\mathcal{C}$. For the fit these two percentages sum to
$100\%$, because at a stationary point of the objective the total abort probability
equals the number of abort labels; this need not hold for the trained Transformer.

\begin{table}[!htbp]
\centering
\caption{The previous-node additive fit against the trained $1/2/384$ Transformer on the same
$2.78$ million supervised positions of one $n=300$ graph.}
\label{tab:addfit}
\begin{tabular}{lcc}
\toprule
Model & \shortstack{Mean $P(\xtok)$ after\\an invalid transition} & \shortstack{Unjustified-abort\\mass fraction} \\
\midrule
Additive \eqref{eq:additive3} & $56.6\%$ & $43.4\%$ \\
Trained, same positions & $99.99\%$ & $0.03\%$ \\
\bottomrule
\end{tabular}
\end{table}

The fit also assigns about $8\%$ abort probability after a valid transition, against
$0.01\%$ for the trained model (Table~\ref{tab:addfit}). Separate additive contributions
therefore do not reproduce the learned rejection quality even with the previous node
available.
% Per-position probabilities and fitted weights are saved with the code.

\paragraph{Planning representations and attention heads.}
\label{sec:planning-probes}
%\siwei{I am wondering whether the title for this subsection is proper.}\shi{I changed a title.}
\siwei{Still think we do not need this subsection, as the section is called "Mechanistic Analysis of Abort Supervision".}
We examine planning information with two probes adapted from \citet{alpine}. Let
$\overline W_V := [W_V^{(1)} \mid \cdots \mid W_V^{(H)}]\WA$ be the value map of the
attention layer when both heads place all their attention on the target position, and
let $e_v$ be the one-hot vector of node $v$. The probes are
\[
\WMp_{(i,\ell)} = \bigl[(\mathrm{FFN}(e_i^\top W_t) + e_i^\top W_t)\WU\bigr]_\ell ,
\qquad
\WVp_{(t,\ell)} = \bigl[(o_t + \mathrm{FFN}(o_t))\WU\bigr]_\ell ,
\]
with $o_t = e_t^\top W_t \overline W_V$. The first probe passes the current-node embedding
through the FFN with its residual connection and probes adjacency information associated
with the current node; the second does the same for the target embedding after the value
map and probes reachability information associated with the target. Here FFN is the trained GELU map, and positional embeddings and layer
normalisations are omitted, following \citet{alpine}. Because the FFN is nonlinear, the
probes are diagnostics of the learned weights, not decompositions of the actual logits;
primes distinguish them from the fitted matrices of \eqref{eq:additive3}.

\begin{figure}[!htbp]
\centering
\includegraphics[width=0.88\textwidth]{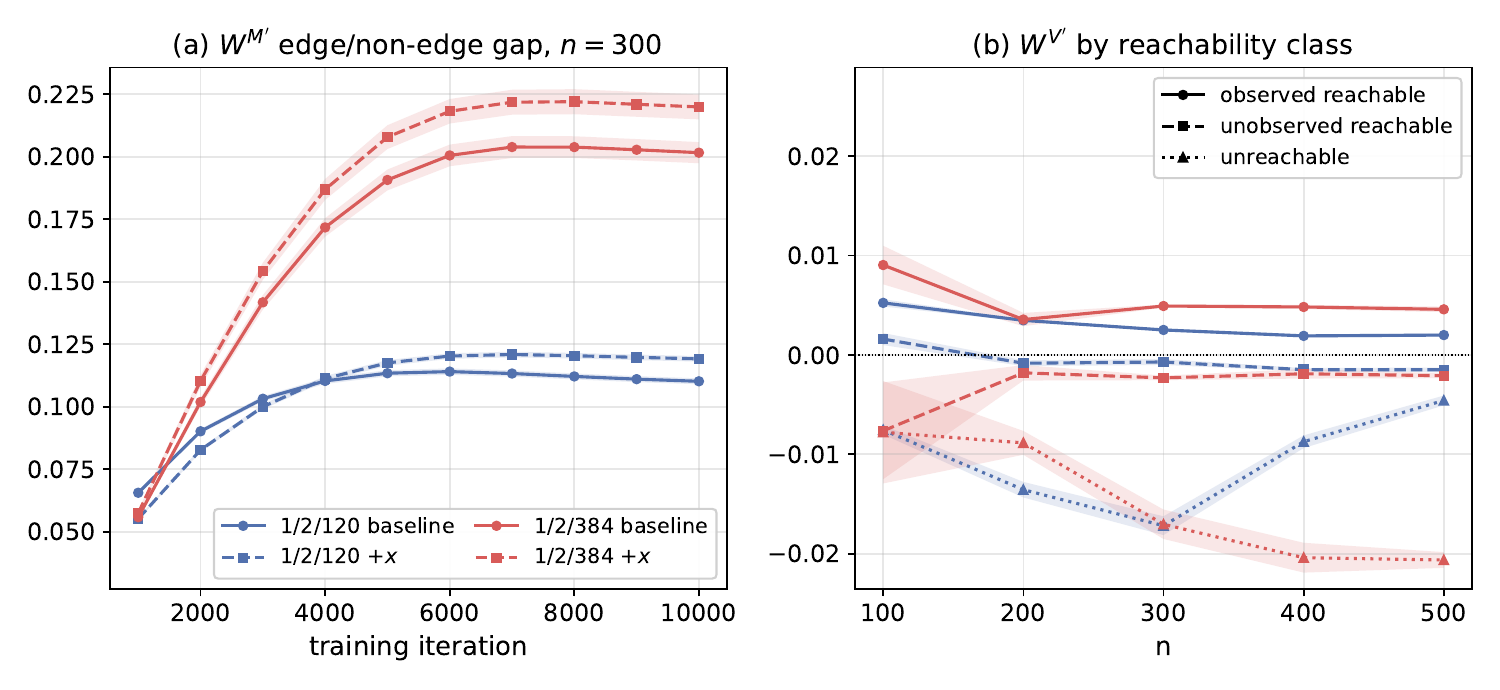}
\caption{Planning representation probes over five seeds; shaded bands are s.e.
(a) Edge--non-edge gap in $\WMp$ at $n=300$.
(b) $\WVp$ by reachability group in baseline models. Architecture colours match (a).}
\label{fig:planning-probes}
\end{figure}

At the end of training, the $\WMp$ edge/non-edge gap of the $+\xtok$ models is slightly
above the baselines' at both sizes, $0.119$ against $0.110$ at $d{=}120$ and $0.220$ against $0.202$ at $d{=}384$
(Figure~\ref{fig:planning-probes}a). The gap compares edges and non-edges among
forward pairs $i<v$ in the graph's node ordering. This is consistent with error
trajectories providing additional adjacency supervision.

For reachability, a pair $(t,u)$ is observed reachable if $\Robs_{(t,u)}=1$, with $\Robs$
as in Remark~\ref{rem:observed-retry}, unobserved reachable if $\Rtrue_{(t,u)}=1$ but
$\Robs_{(t,u)}=0$, and unreachable if $\Rtrue_{(t,u)}=0$. Figure~\ref{fig:planning-probes}b
shows mean $\WVp$ values over $u<t$ for these groups in the two-head baseline models.
The observed-reachable minus unobserved-reachable gap
is positive across all three architectures and five graph sizes, and the
unobserved-reachable minus unreachable gap is larger in $13$ of these $15$
configurations, so $\WVp$ carries some information about true reachability beyond
the observed pairs, with additional separation for observed pairs.

%
%We measure the composition of observed path segments by the \emph{reachability degree}:
%the minimum number of edges from $s$ to a node $v$ with $\Robs_{(t,v)}=1$.
%Path generation remains poor at degree $\geq2$ (Appendix~\ref{app:diagnostics}).

At
$1/2/384$ one head attends to the target and the other to the current position in all
five seeds, for both baseline and $+\xtok$ models (Figure~\ref{fig:heads}); which head
takes which role varies between runs.

\subsection{Additional Graph Planning Results}
\label{app:controls}

\paragraph{Single-shot validity and model size.}
Abort supervision leaves single-shot validity unchanged at $n=100$ for all three
architectures. It lowers it for the $d{=}120$ models by $2.0\pm0.2$ and $2.5\pm0.3$
percentage points at $n=200$ (one and two heads) and by $11.6\pm0.3$ and $11.3\pm0.4$
points at $n=500$; at $d{=}384$ the change is within the error bars at $n=200$ and between
$-0.9\pm0.1$ and $-2.8\pm0.0$ points from $n=300$ on (Figure~\ref{fig:gain}a). The cost is
larger for the $d{=}120$ models, whose abort precision is also lower ($68$--$94\%$, against
$94$--$99\%$ at $d{=}384$). The difference is associated with embedding size rather than head count:
$1/2/120$ tracks $1/1/120$ within about one point at the same parameter count, which is
why \S\ref{sec:empirical} uses the two-head models.

\paragraph{Matched correct-trajectory supervision.}
\label{app:budget}
Because half of the trajectories in a $+\xtok$ batch carry a single label, the $+\xtok$
model receives half the baseline's correct-trajectory supervision per update
(\S\ref{par:fairness}). We therefore also train both $1/2/384$ models for $20{,}000$
updates on the $n=300$ graph of Figure~\ref{fig:heads}, after which the $+\xtok$ model has
seen the correct-trajectory supervision the baseline receives in $10{,}000$. Harness
validity in percent on that graph:
\begin{center}
\begin{tabular}{lccc}
\toprule
updates & baseline $+$ retry & $+\xtok$ $+$ retry & gain \\
\midrule
$10{,}000$ (the setting used throughout) & $97.6$ & $99.6$ & $+2.0$ \\
$20{,}000$ & $96.9$ & $99.6$ & $+2.7$ \\
\bottomrule
\end{tabular}
\end{center}
At matched correct-trajectory exposure, the $20{,}000$-update $+\xtok$ model exceeds
the $10{,}000$-update baseline by $2.0$ percentage points. The longer-trained baseline
is slightly worse, its invalid-transition rate rising from $2.4\%$ to $3.1\%$.

\subsection{Additional Tasks and Method Comparisons}
\label{app:additional}

\paragraph{A pretrained model on Blocksworld.}
\label{app:blocksworld}

The three domains above use one-layer models trained from scratch. We also evaluate
abort supervision on a pretrained language model, Qwen3-0.6B-Base, on Blocksworld
with $8$ or $9$ blocks. A move takes an uncovered block and places it on the table
or on another uncovered block. The prompt gives the initial and goal configurations as stacks; the model writes
one move per line (\texttt{move A onto B}, \texttt{move A onto table}) and then
\texttt{done}, and never restates the state. Each instance is an initial--goal pair whose
shortest solution, found by breadth-first search from the goal, has $4$--$15$ moves. Of
$300$ goals ($150$ per block count), $255$ are training goals with $40$ initial states
each, giving $10{,}200$ plans; $45$ are held-out goals with $60$ initial states each, and
evaluation uses $2{,}000$ of these $2{,}700$ instances. Error trajectories take a valid prefix of a training plan, append
one invalid move (moving a buried block, moving onto a buried block, or a no-op) and a new
token \texttt{<|abort|>} playing the role of \xtok, with the loss on that token only;
$\rho = 8$. Both models are fine-tuned for $3$ epochs (learning rate $2\times10^{-5}$,
batch size $64$) on one NVIDIA A100 GPU; because the $+\xtok$ training set is nine times
as large, it receives nine times as many updates, and a second baseline trained for $27$
epochs matches its update count. The harness draws at most $K=10$ plans per instance at
temperature $1$ and accepts the first that does not abort and ends with \texttt{done}; it
does not check moves or the final state, and validity is scored by simulation.

\begin{figure}[!htbp]
\centering
\includegraphics[width=\textwidth]{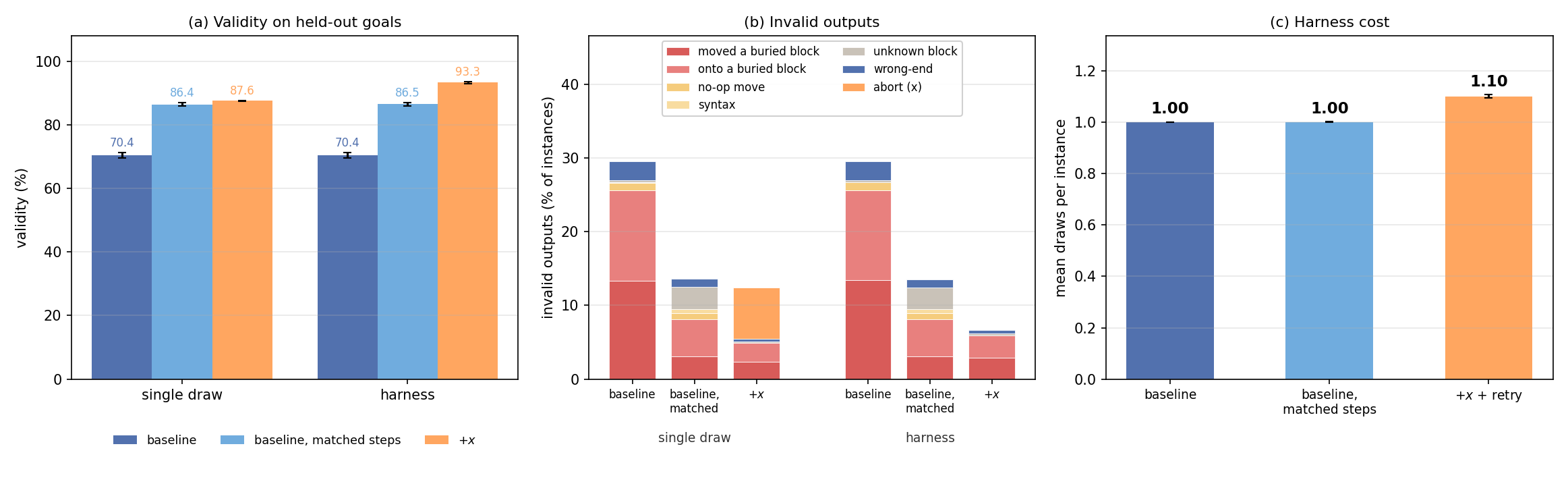}
\caption{Blocksworld, Qwen3-0.6B-Base; mean $\pm$ s.e.\ over three seeds. (a) Validity for a single
draw and under the harness. (b) Invalid outputs for a single draw and for the harness output.
(c) Mean draws per instance.}
\label{fig:blocksworld}
\end{figure}

Retrying on the abort raises harness validity from $70.4 \pm 0.9\%$ to
$93.3 \pm 0.3\%$, a gain of $+22.9 \pm 1.0$ percentage points at $1.10$ draws per
instance, and of $+6.8 \pm 0.6$ points over the baseline trained for the same number of
updates. Illegal moves account for most invalid outputs in every arm: they appear in $26.7\%$ of
the outputs returned by the baseline harness, $8.9\%$ for the matched-steps baseline and
$5.9\%$ for the $+\xtok$ harness (Figure~\ref{fig:blocksworld}b). At the coverage of the
$+\xtok$ model, $93.1\%$ of instances, its selective accuracy is $94.1\%$, against $75.1\%$
for a mean-log-probability threshold on the baseline and $88.4\%$ on the matched-steps
baseline.

\paragraph{Self-Backtracking on Countdown.}
\label{app:selfback}
We adapt Self-Backtracking~\citep{selfback} to the same error trajectories as $+\xtok$.
Its loss supervises the correct prefix and a backtrack token after the invalid step,
masking only that step. Training otherwise matches the Countdown runs of
\S\ref{sec:endtoend}: the same $1/2/384$ model, split, error trajectories at $\rho=1$,
$10{,}000$ updates, optimiser and five seeds, and the backtrack token uses the same
vocabulary entry as the abort token. The baseline and $+\xtok$ models are the $\rho=1$
models of \S\ref{sec:endtoend}, evaluated on the same $3{,}000$ test instances with fresh
samples, so their single-shot validity differs slightly from the values there.

At inference, Self-Backtracking first samples $N$ trajectories and keeps completed
outputs without a backtrack token as candidates. For outputs that emit the token,
it removes the last equation and selects up to $\lfloor\sqrt N\rfloor$ distinct
shortened prefixes. It then samples $N$ continuations from each selected prefix.
This is one round of backtracking, denoted $b=1$. The likelihood selection rule
returns the candidate with the highest mean token log-probability. We also evaluate
the harness rule of \S\ref{sec:endtoend}, which selects the first candidate whose
last written value is the target.

For each instance, the output-token count used by Self-Backtracking at a given $N$
sets a hard budget for baseline and $+\xtok$ restart. A final restart draw that
would exceed this budget is truncated and discarded. We also exchange the inference
procedures to separate their effect from the training loss: the Self-Backtracking
model uses restart under the same token budget, and the $+\xtok$ model uses
backtracking with the same $N$. The latter has similar, but not identical, token
counts. All combinations are evaluated under both selection rules; the resource
comparison concerns generated output tokens, not wall-clock time.

\begin{figure}[!htbp]
\centering
\includegraphics[width=\textwidth]{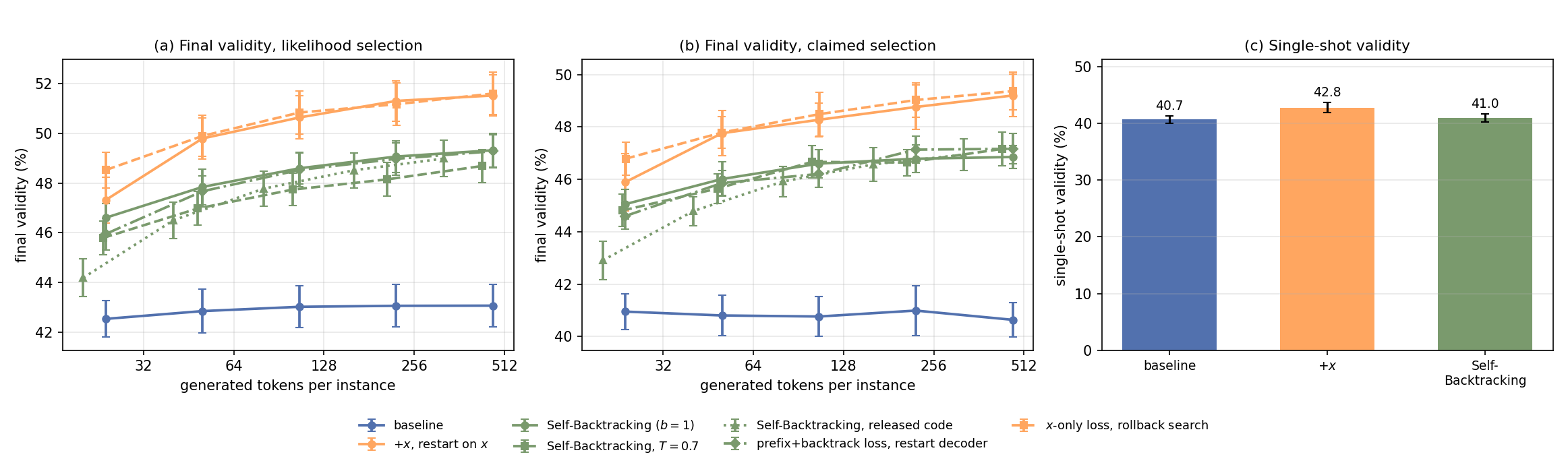}
\caption{Countdown, five seeds, mean $\pm$ s.e. (a), (b) Final validity against output tokens per
instance under the likelihood and harness selection rules; budgets are set by
Self-Backtracking at $N\in\{2,4,8,16,32\}$. Dashed orange and dash-dotted green exchange
the inference procedures. (c) Single-shot validity.}
\label{fig:selfback}
\end{figure}

Figure~\ref{fig:selfback} shows the results. Single-shot validity rises by
$+2.1\pm0.5$ percentage points for $+\xtok$ and by $+0.3\pm0.7$ for Self-Backtracking. Under the likelihood rule at $N=32$ (about $470$
tokens per instance) the baseline reaches $43.1\pm0.8\%$, Self-Backtracking
$49.3\pm0.7\%$ and $+\xtok$ $51.5\pm0.8\%$. The paired difference between $+\xtok$ and
Self-Backtracking is $+2.0$ to $+2.2$ points at every budget from $N=4$ on and positive on
every seed, and $+1.7$ to $+2.3$ points under the harness rule ($+0.7\pm0.7$ and
$+0.8\pm0.9$ at $N=2$). Exchanging inference procedures suggests that the loss accounts
for most of the difference in this setting: at $N=32$, the $+\xtok$ model with
backtracking reaches $51.6\pm0.9\%$, and the Self-Backtracking model with restart
reaches $49.3\pm0.7\%$, close to their results with the original procedures. One round
of backtracking adds $0.3$ to $1.6$ points over selecting from the same initial samples
without backtracking ($b=0$). We also evaluate the released-code variant, which expands
only one shortened prefix and does not expand it if no equations remain, and a
temperature-$0.7$ variant. Both are shown in the figure and perform at or below the
$\lfloor\sqrt N\rfloor$ rule at temperature $1$ under likelihood selection.

\end{document}